\documentclass[11pt,a4paper]{article}
\usepackage[hyperref]{emnlp2020}
\usepackage{times}
\usepackage{latexsym}

\aclfinalcopy 
\def\aclpaperid{2705} 

\usepackage{amsmath}
\usepackage{amssymb}
\usepackage{amsfonts}
\usepackage{bbm}
\usepackage{booktabs}
\usepackage{diagbox}
\usepackage{graphicx}
\usepackage{makecell}
\usepackage{microtype}
\usepackage{multirow}
\usepackage{paralist}
\usepackage{subcaption}
\usepackage{xcolor}
\usepackage{xspace}

\usepackage[normalem]{ulem}

\newcommand{\seq}[1]{\boldsymbol{#1}}
\newcommand{\gold}{y^*}

\newcommand{\correctness}{{$\mathbf{correctness}$}\xspace}
\newcommand{\confidence}{{$\mathbf{confidence}$}\xspace}
\newcommand{\variance}{{$\mathbf{variability}$}\xspace}

\newcommand{\bert}{\textsc{BERT}\xspace}
\newcommand{\roberta}{\textsc{RoBERTa}\xspace}

\newcommand{\adversarial}{\textsl{{Adversarial SQuAD}}\xspace}
\newcommand{\diagnostics}{\textsl{{NLI Diagnostics}}\xspace}

\newcommand{\mnli}{{\textsl{MultiNLI}}\xspace}
\newcommand{\qnli}{{\textsl{QNLI}}\xspace}
\newcommand{\snli}{\textsl{{SNLI}}\xspace}
\newcommand{\squad}{{\textsl{SQuAD}}\xspace}
\newcommand{\winogrande}{\textsl{{WinoGrande}}\xspace}
\newcommand{\wsc}{\textsl{{WSC}}\xspace}

\newcommand{\draftonly}[1]{#1}
\renewcommand{\draftonly}[1]{} 
\newcommand{\draftcomment}[3]{\draftonly{\textcolor{#2}{[#3]{$_{\textsc{#1}}$}}}}
\newcommand{\todo}[1]{\draftcomment{TODO}{red}{#1}}
\newcommand{\swabha}[1]{\draftcomment{ss}{teal}{#1}}

\newcommand{\nascomment}[1]{\draftcomment{nas}{cyan}{#1}}

\newcommand{\com}[1]{}
\newcommand{\tabref}[1]{Tab.~\ref{tab:#1}}
\newcommand{\figref}[1]{Fig.~\ref{fig:#1}}

\newcommand{\easy}[0]{\textsl{easy-to-learn}\xspace}
\newcommand{\Easy}[0]{\textsl{Easy-to-Learn}\xspace}
\newcommand{\noisy}[0]{\textsl{hard-to-learn}\xspace}
\newcommand{\Noisy}[0]{\textsl{Hard-to-learn}\xspace}
\newcommand{\challenging}[0]{\textsl{ambiguous}\xspace}

\title{Dataset Cartography:\\Mapping and Diagnosing Datasets with Training Dynamics}

\author{
    Swabha Swayamdipta$^\dagger$ \quad
	Roy Schwartz$^{\ddagger}$\thanks{~~Work done at the Allen Institute for AI.} \quad
	Nicholas Lourie$^{\dagger}$ \quad \\
	\bf Yizhong Wang$^\diamondsuit$ \quad
	Hannaneh Hajishirzi$^{\dagger\diamondsuit}$ \quad
	Noah A. Smith$^{\dagger\diamondsuit}$ \quad 
	Yejin Choi$^{\dagger\diamondsuit}$ \\\\
	$^\dagger$Allen Institute for Artificial Intelligence, Seattle, WA, USA \\
	$^\ddagger$The Hebrew University of Jerusalem, Israel\\
	$^\diamondsuit$Paul G. Allen School of Computer Science \& Engineering, University of Washington, Seattle, WA, USA \\
	{\tt \{swabhas,nicholasl\}@allenai.org 
	roys@cs.huji.ac.il} \\ 
	{\tt \{yizhongw,hannaneh,noah,yejin\}@cs.washington.edu}
}
\date{}

\begin{document}
\maketitle

\begin{abstract}
Large datasets have become commonplace in NLP research.
However, the increased emphasis on data quantity has made it challenging to assess the quality of data.
We introduce \textit{Data Maps}---a model-based tool to characterize and diagnose datasets.
We leverage a largely ignored source of information: the behavior of the model on individual instances during training (\emph{training dynamics}) for building data maps. 
This yields two intuitive measures for each example---the model's confidence in the \textit{true} class, and the variability of this confidence across epochs---obtained in a single run of training.
Experiments across four datasets show that these model-dependent measures reveal three distinct regions in the data map, each with pronounced characteristics.
First, our data maps show the presence of \textit{ambiguous} regions with respect to the model, which contribute the most towards out-of-distribution generalization.
Second, the most populous regions in the data are \textit{easy to learn} for the model, and play an important role in model optimization.
Finally, data maps uncover a region with instances that the model finds \textit{hard to learn}; these often correspond to labeling errors.
Our results indicate that a shift in focus from quantity to quality of data could lead to robust models and improved out-of-distribution generalization.
\end{abstract}

\section{Introduction}
\label{sec:intro}

\begin{figure}[t]
     \centering
     \includegraphics[width=\columnwidth]{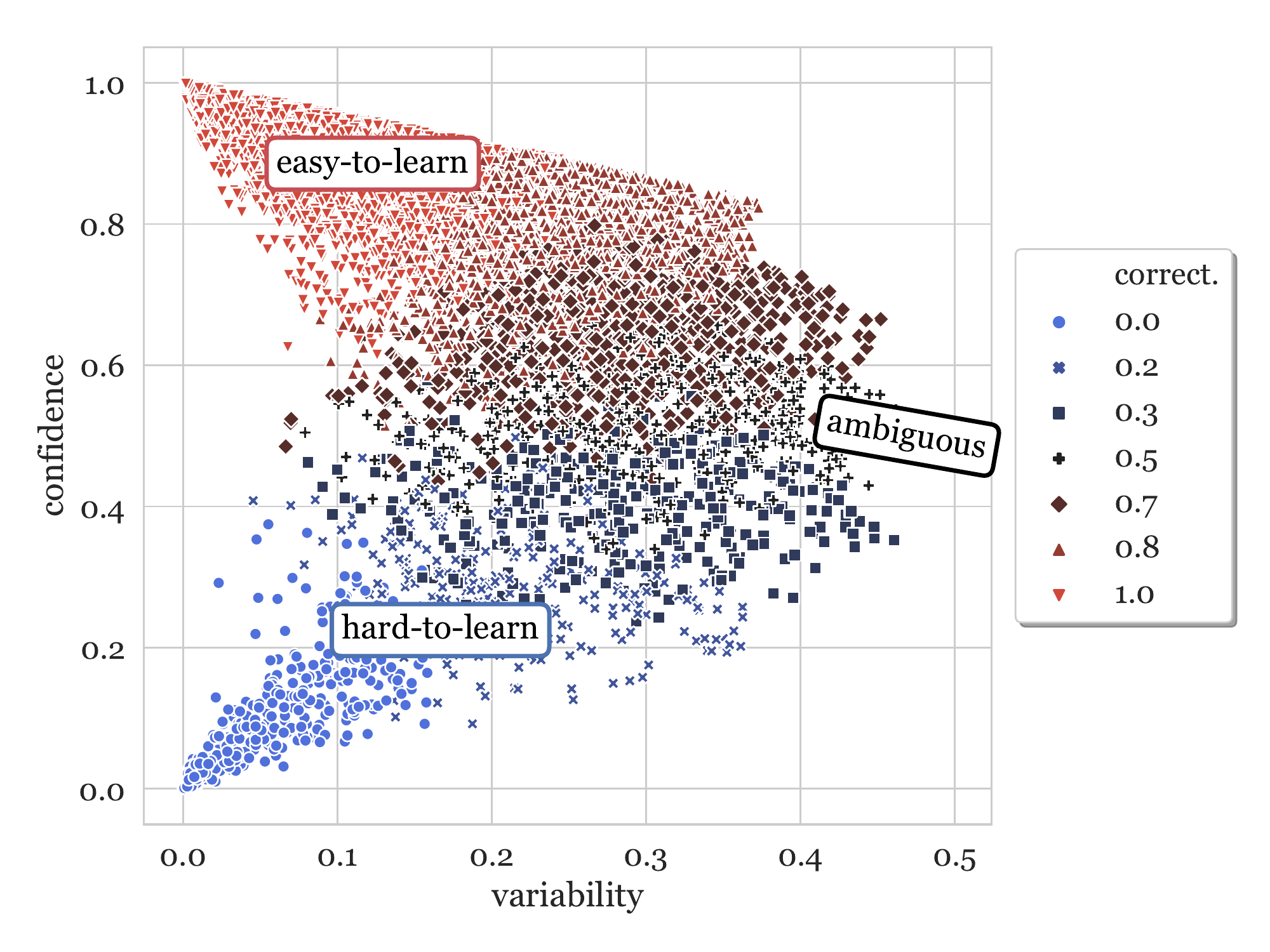}
     \caption{ 
     Data map for \snli train set, based on a \roberta-large classifier. 
    The $x$-axis shows \variance and $y$-axis, the \confidence; the colors/shapes indicate \correctness.
     The top-left corner of the data map (low \variance, high \confidence) corresponds to \textbf{\easy} examples, the bottom-left corner (low \variance, low \confidence) corresponds to \textbf{\noisy} examples, and examples on the right (with high \variance) are \textbf{\challenging}; all definitions are with respect to the \roberta-large model.
     The modal group in the data is formed by the \easy regions. 
     For clarity we only plot 25K random samples from the \snli train set. \figref{snli2} in App.~\S\ref{app:all_data_maps} shows the same map in greater relief.
     } 
     \label{fig:snli}
 \end{figure}

The creation of large labeled datasets has fueled the advance of AI \cite{Russakovsky2015Imagenet,agrawal2015vqa}
and NLP in particular \cite{bowman-etal-2015-large,Rajpurkar2016SQuAD10}.
The common belief is that the more abundant the labeled data, the higher the likelihood of learning diverse phenomena, which in turn leads to models that generalize well.
In practice, however, out-of-distribution (OOD) generalization remains a challenge \cite{yogatama2019learning, linzen2020can}; and, while recent large pretrained language models help, they fail to close this gap \cite{hendrycks2020pretrained}.
This urges a closer look at datasets, where not all examples might contribute equally towards learning \cite{vodrahalli2018training}.
However, the scale of data can make this assessment challenging.
How can we \textit{automatically} characterize data instances with respect to their role in achieving good performance in- and out-of- distribution?
Answering this question may take us a step closer to bridging the gap between dataset collection and broader task objectives.  

Drawing analogies from cartography, we propose to find coordinates for instances within the broader trends of a dataset.
We introduce \emph{data maps}: a model-based tool for contextualizing examples in a dataset.
We construct coordinates for data maps by leveraging \textit{training dynamics}---the behavior of a model as training progresses.
We consider the mean and standard deviation of the \textit{gold label} probabilities, 
predicted for each example across training epochs; these are referred to as \confidence and \variance, respectively (\S\ref{sec:mapping}).

\figref{snli} shows the data map for the \snli dataset \cite{bowman-etal-2015-large} constructed using the \roberta-large model \cite{Liu2019RoBERTaAR}.
The map reveals three distinct regions in the dataset: a region with instances whose true class probabilities fluctuate frequently during training (high \variance), and are hence \textit{\challenging} for the model; a region with \textit{\easy} instances that the model predicts correctly and consistently (high \confidence, low \variance); and a region with \textit{\noisy} instances with low \confidence, low \variance, many of which we find are \textit{mis}labeled during annotation .\footnote{All terms are defined with respect to the model.}
Similar regions are observed across three other datasets: \mnli \cite{williams-etal-2018-broad}, \winogrande \cite{sakaguchi2019winogrande} and \squad \cite{Rajpurkar2016SQuAD10}, with respect to respective \roberta-large classifiers.

We further investigate the above regions by training models exclusively on examples from each region (\S\ref{sec:selection}).
Training on \challenging instances promotes generalization to OOD test sets, with little or no effect on in-distribution (ID) performance.\footnote{We define out-of-distribution (OOD) test sets as those which are collected independently of the original dataset, and ID test sets as those which are sampled from it.}
Our data maps also reveal that  datasets contain a majority of \easy instances, which are not as critical for ID or OOD performance, but without any such instances, training could fail to converge (\S\ref{sec:easy}).
In \S\ref{sec:noise}, we show that \noisy instances frequently correspond to labeling errors.
Lastly, we discuss connections between our measures and uncertainty measures (\S\ref{sec:analysis}).

Our findings indicate that data maps could serve as effective tools to diagnose large datasets, at the reasonable cost of training a model on them.
Locating different regions within the data might pave the way for constructing higher quality datasets., and ultimately models that generalize better.
Our code and higher resolution visualizations are publicly available.\footnote{\url{https://github.com/allenai/cartography}}

\section{Mapping Datasets with Training Dynamics}
\label{sec:mapping}

Our goal is to construct \textit{Data Maps} for datasets to help visualize a dataset with respect to a model, as well as understand the contributions of different groups of instances towards that model's learning.
Intuitively, instances that a model always predicts correctly are different from those it almost never does, or those on which it vacillates.
For building such maps, each instance in the dataset must be contextualized in the larger set.
We consider one contextualization approach, based on statistics arising from the behavior of the training procedure across time, or the ``training dynamics''.
We formally define our notations (\S\ref{sec:training_dynamics}) and describe our data maps (\S\ref{sec:data_maps}).

\begin{figure*}[t]
     \centering
     \includegraphics[width=\textwidth]{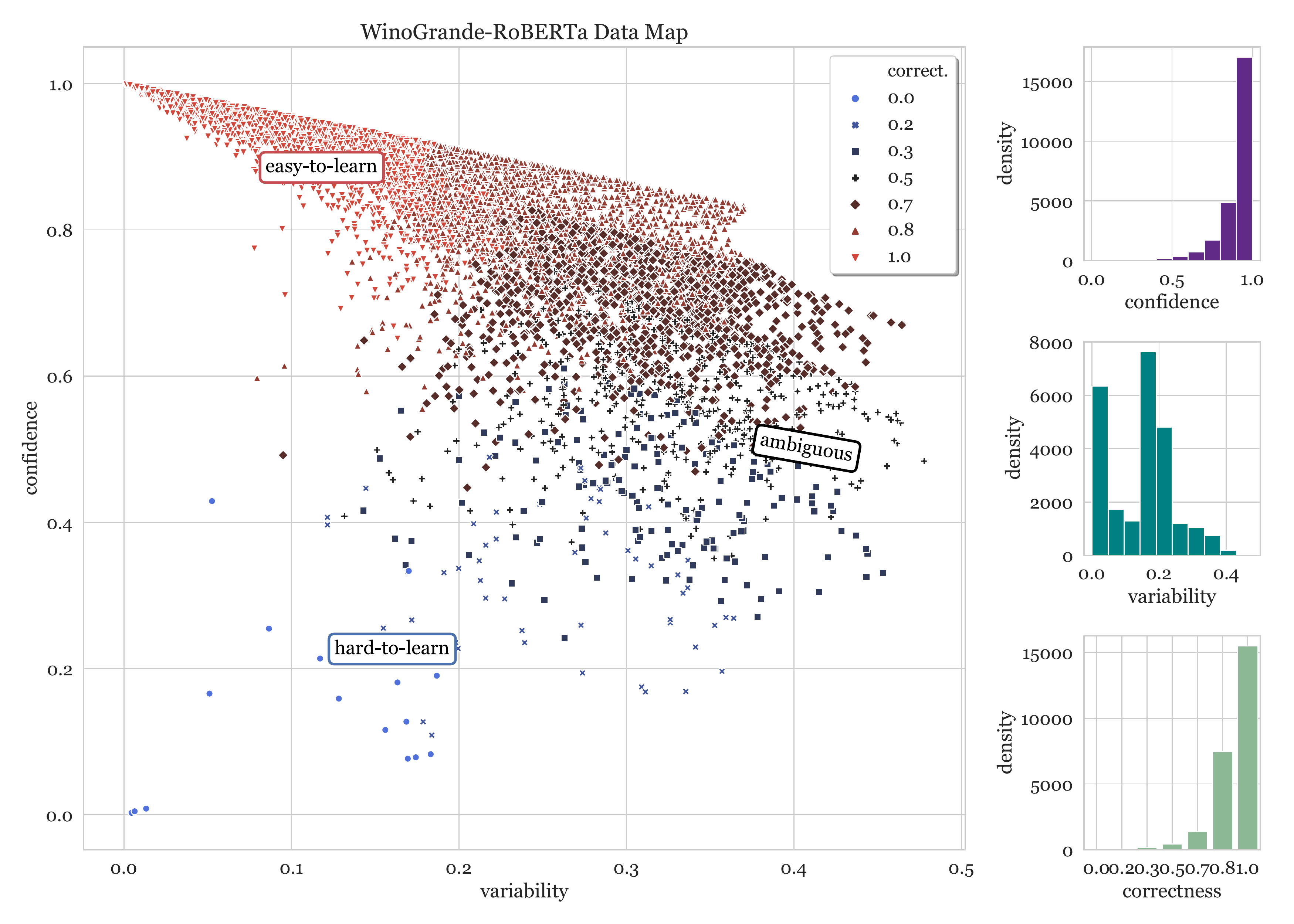}
     \caption{ 
     Data map for the \winogrande \cite{sakaguchi2019winogrande} train set, based on a \roberta-large classifier, with the same axes as \figref{snli}.
     Density plots for the three different measures based on training dynamics are shown towards the right.
     \Noisy regions have lower density in \winogrande, compared to \snli, perhaps as a result of a rigorous validation of collected annotations. 
     However, manual errors remain, which we showcase in \tabref{qualitative} as well as in Section \S\ref{sec:noise}.
     The plot shows only 25K train examples for clarity, and is best viewed enlarged.
     } 
     \label{fig:winogrande}
 \end{figure*}

\subsection{Training Dynamics}
\label{sec:training_dynamics}

Consider a training dataset of size $N$, $\mathcal{D} = \{(\seq{x}, \gold)_i\}_{i=1}^{N}$ where the $i$th instance consists of the observation, $\seq{x}_i$ and its true label under the task, $\gold_i$.
Our method assumes a particular model (family) whose parameters are selected to minimize empirical risk using a particular algorithm.\footnote{In this paper, the model is \roberta \cite{Liu2019RoBERTaAR}, currently established as a strong performer across many tasks.}
We assume the model defines a probability distribution over labels given an observation.
We assume a stochastic gradient-based optimization procedure is used, with training instances randomly ordered at each epoch, across $E$ epochs.

The training dynamics of instance $i$ are defined as statistics calculated across the $E$ epochs. 
The values of these measures then serve as coordinates in our map. 
The first measure aims to capture how confidently the learner assigns the \textit{true label} to the  observation, based on its probability distribution.
We define \confidence as the mean model probability of the true label ($\gold_i$) across epochs:
\begin{align*}
    \hat{\mu}_i &= \frac{1}{E} \sum_{e=1}^E p_{\boldsymbol{\theta}^{(e)}} (\gold_i \mid \boldsymbol{x}_i)
\end{align*}
where $p_{\boldsymbol{\theta}^{(e)}}$ denotes the model's probability with parameters $\boldsymbol{\theta}^{(e)}$ at the end of the $e^\text{th}$ epoch.\footnote{Note that $\hat{\mu}_i$ is with respect to the true label $\gold_i$, not the probability assigned to the model's highest-scoring label (as used in active learning, for example).}
In some cases we also consider a coarser, and perhaps more intuitive statistic, the fraction of times the model correctly labels $\boldsymbol{x}_i$ across epochs, named \correctness; this score only has $1 + E$ possible values.  
Intuitively, a high-\confidence instance is ``easier'' for the given learner.

Lastly, we also consider \variance, which measures the spread of $p_{\boldsymbol{\theta}^{(e)}}(\gold_i \mid \boldsymbol{x}_i)$ across epochs, using the standard deviation:
\begin{align*}
    \hat{\sigma}_i =& \sqrt{\frac{\sum_{e =1}^E \left (p_{\boldsymbol{\theta}^{(e)}}(\gold_i \mid \seq{x}_i) - \hat{\mu}_i \right )^2}{E}}
\end{align*}
Note that \variance also depends on the gold label, $\gold_i$.  
A given instance to which the model assigns the same label consistently (whether accurately or not) will have low \variance; one which the model is indecisive about across training, will have high \variance.

Finally, we observe that \confidence and \variance are fairly stable across different parameter initializations.\footnote{The average Pearson correlation coefficient between five random seeds' resulting training runs is 0.75 or higher (for both measures, on \winogrande).}
Training dynamics can be computed at different granularities, such as steps vs.~epochs; see App.~\ref{sec:efficient_training_dynamics}.

\subsection{Data Maps}
\label{sec:data_maps}

We construct data maps for four large datasets: 
\winogrande \cite{sakaguchi2019winogrande}---a cloze-style task for commonsense reasoning, two NLI datasets (\snli;  \citealp{bowman-etal-2015-large}; and \mnli; \citealp{williams-etal-2018-broad}), and \qnli, which is a sentence-level question answering task derived from \squad \cite{Rajpurkar2016SQuAD10}.
All data maps are built with models based on \roberta-large architectures.
Details on the model and datasets can be found in App.~\S\ref{app:data} and \S\ref{app:experiments}.

\figref{snli} presents the data map for the \snli dataset.
As is evident, the data follows a bell-shaped curve with respect to \confidence and \variance; \correctness further determines discrete regions therein.
The vast majority of instances belong to the high \confidence and low \variance region of the map (\figref{snli}, top-left). 
The model \textit{consistently} predicts such instances correctly with \textit{high} confidence; thus, we refer to them as \textbf{\easy} (for the model).
A second, smaller group is formed by instances with low \variance and low \confidence (\figref{snli}, bottom-left corner). 
Since such instances are seldom predicted correctly during training, we refer to them as \textbf{\noisy} (for the model).
The third notable group contains \challenging examples, or those with high \variance (\figref{snli}, right-hand side); the model tends to be indecisive about these instances, such that they may or may not correspond to high \confidence or \correctness. 
We refer to such instances as \textbf{\challenging} (to the model).

\begin{table}[t]
\centering
\scriptsize
\begin{tabular}{p{.2cm}p{3.6cm}cc}
\toprule 
  & \textbf{Instance} & \textbf{Option1} & \textbf{Option2} \\
\midrule 
\multirow{5}{*}{\rotatebox[origin=c]{90}{\easy}}
& The man chose to buy the roses instead of the carnations because the \_\_ were more beautiful.
& \colorbox{orange!30!}{roses*} 
& carnations
\\\cmidrule[0.01em](lr){2-4}
& We enjoyed the meeting tonight but not the play as the \_\_ was rather dull.
& meeting
& \colorbox{orange!30!}{play*} 
\\\midrule[0.03em]
\multirow{11}{*}{\rotatebox[origin=c]{90}{\noisy}}
& Jason got into a deep financial hole, unlike Joel, because \_\_ managed their fortune poorly.
& \colorbox{blue!30!}{Jason$^+$} 
& \colorbox{orange!30!}{Joel*} 
\\\cmidrule[0.01em](lr){2-4}
& In the mornings, Aaron can hit the snooze button a lot, and Samuel can't. \_\_ has to be at work at 10 am.
& \colorbox{orange!30!}{Aaron*}
& \colorbox{green!30!}{Samuel$^-$} 
\\\cmidrule[0.01em](lr){2-4}
& Amy's handwriting was meticulous, while Cynthia's handwriting was often sloppy, because \_\_ was careless about their work.
& \colorbox{orange!30!}{Amy*} 
& \colorbox{blue!30!}{Cynthia$^+$} 
\\\midrule[0.03em]
\multirow{6}{*}{\rotatebox[origin=c]{90}{\challenging}}
& The dog ran up to Leslie and away from Lawrence because \_\_ had soap for the dog to take a bath.
& \colorbox{green!30!}{Leslie$^-$} 
& \colorbox{orange!30!}{Lawrence*}
\\\cmidrule[0.01em](lr){2-4}
& Kayla dated many more people at once than Betty, because \_\_ was in an exclusive relationship.	
& \colorbox{orange!30!}{Kayla*}
& \colorbox{blue!30!}{Betty$^+$}
 \\
\bottomrule
\end{tabular}
\caption{
Examples from the \winogrande train set from different regions in the data map, with \colorbox{orange!30!}{gold standard*} labels. 
Our best assessment of the \colorbox{blue!30!}{correct$^+$} and \colorbox{green!30!}{equally plausible$^-$} labels are highlighted.
} 
\label{tab:qualitative}
\end{table}

\figref{winogrande} shows the data map for \winogrande, which exhibits high structural similarity to the \snli data map (\figref{snli}).
The most remarkable difference between the maps is in the density of the \noisy region, which is much lower for \winogrande, as is evident from the histograms below.
One explanation for this might be that \winogrande labels were rigorously validated post annotation.
App.~\S\ref{app:all_data_maps} includes data maps for all four datasets, with respect to \roberta-large, in greater relief.

Different model architectures trained on a given dataset could be effectively compared using data maps, as an alternative to standard quantitative evaluation methods.
App.~\S\ref{app:all_data_maps} includes data maps for \winogrande (\figref{bert_winogrande}) and \snli (\figref{lstm_bow_snli} and \figref{bert_esim_snli}) based on other (somewhat weaker) architectures.
While data maps based on similar architectures have similar appearance, the regions to which a given instance belongs might vary.
Data maps for weaker architectures still display similar regions, but the regions are not as distinct as those in \roberta based data maps.

\tabref{qualitative} shows examples from \winogrande belonging to the different regions defined above.
\easy examples are straightforward for the model, as well as for humans.
In contrast, most \noisy and some \challenging examples could be challenging for humans (see green highlights in \tabref{qualitative}), which might explain why the model shows lower \confidence on them.
These categories could be harder for models either because of labeling errors (blue highlights) or simply because the model is indecisive about the correct label.
See App.~\S\ref{app:qualitative} for similar examples from \snli.

The next four sections include a diagnosis of the different data regions defined above.
The effect of training models on each region on both in- and out-of-distribution performance is studied in \S\ref{sec:selection}.
The effect of selecting decreasing amounts of data is discussed in \S\ref{sec:easy}.
We investigate the presence of mislabeled instances in the \noisy regions of the data maps in \S\ref{sec:noise}.
Lastly, we demonstrate connections between training dynamics measures and measures of uncertainty in \S\ref{sec:analysis}.

\section{Data Selection using Data Maps}
\label{sec:selection}

Data maps reveal distinct regions in datasets; 
it is natural to wonder what roles do instances from different regions play in learning and generalization.
We answer this empirically by training models \textit{exclusively} on instances selected from distinct regions, followed by standard in-distribution (ID), as well as out-of-distribution (OOD) evaluation.

Our strategy is straightforward---we train 
the model from scratch on a subset of the training data selected by ranking instances based on the different training dynamics measures.\footnote{Hyperparameters are also tuned from scratch (App.~\S\ref{app:experiments}).}
We hypothesize that \challenging and \noisy regions could be the most informative for learning, since these examples are the most challenging for the model \cite{shrivastava2016training}.
We compare these two settings to \roberta-large models trained on data subsets, selected using several other methods. 
All subsets considered contain 33\% of the training data (to control for the effect of train data size on performance).

\paragraph{Baselines}
\label{sec:baselines}

The two most natural baselines are those where all data is used (\textbf{{100\% train}}), and where a 33\% random sample is used (\textbf{{random}}).
Our second set of baselines considers subsets which are the most \easy for the model (\textbf{{high-\confidence}}), and those that the model is most decisive about (\textbf{{low-\variance}}), which comprises a mixture of \easy and \noisy examples.
We also consider baselines based on \textbf{{high-}} and \textbf{\textbf{low-\correctness}}. 
Finally, we also compare with our implementation of the following methods for data selection from prior work (discussed in \S\ref{sec:related}):
\textbf{{forgetting}} \cite{toneva2018empirical},
\textbf{{AFLite}} \cite{lebras2020adversarial},
\textbf{{AL-uncertainty}} \cite{joshi-2009-multi}, and 
\textbf{{AL-greedyK}} \cite{sener2018active}.

\begin{table}
\setlength{\tabcolsep}{3pt}
\small
\centering
\begin{tabular}{llcc}
\toprule
\cmidrule(lr){3-3}  \cmidrule(lr){4-4} 
& & \textsc{WinoG.} Val. (ID) &  \wsc  (OOD)  \\
\midrule
 & {100\% train}        & 79.7$_{0.2}$ & 86.0$_{0.1}$ \\
\midrule[0.03em]
\multirow{11}{*}{\rotatebox[origin=c]{90}{{33\% train}}} & {random}              & 73.3$_{1.3}$ & 85.6$_{0.4}$ \\ 
\cmidrule[0.03em]{2-4}
& {high-}\correctness   & 70.8$_{0.6}$ & 84.1$_{0.4}$ \\
& {high-}\confidence    & 69.4$_{0.5}$ & 83.9$_{0.5}$ \\
& {low-}\variance       & 70.1$_{1.0}$ & 83.7$_{1.4}$ \\
\cmidrule[0.03em]{2-4}
& {forgetting}          & 75.5$_{1.3}$ & 84.8$_{0.7}$ \\
& {AL-uncertainty}      & 75.7$_{0.8}$ & 85.7$_{0.8}$ \\
& {AL-greedyK }         & 74.2$_{0.4}$ & 86.5$_{0.5}$ \\
& {AFLite}              & 76.8$_{0.8}$ & 86.6$_{0.6}$ \\
\cmidrule[0.03em]{2-4}
& {low-}\correctness    & 78.2$_{0.6}$ & 86.3$_{0.6}$  \\
& \noisy                & 77.9$_{1.3}$ & 87.2$_{0.7}$ \\
& \challenging          & \textbf{78.7}$_{0.4}$ & \textbf{87.6}$_{0.6}$ \\
\bottomrule
\end{tabular}
\caption{
ID and OOD accuracies for \roberta-large models trained on different selections of \winogrande.
Reported values are averaged over 3 random seeds, with s.d.~reported as a subscript.
Selection of 33\% training instances with highest \variance (\challenging) achieves the best OOD performance, outperforming all other baselines from this work, as well as prior work.
}
\label{tab:winogrande}
\end{table}
\begin{table*}
\footnotesize
\centering
\begin{tabular}{llcccccc|ccccccc}
\toprule
\multicolumn{2}{c}{}  & \multicolumn{6}{c}{\snli}  & \multicolumn{7}{c}{\mnli} \\
\cmidrule(lr){3-8} \cmidrule{9-15} \multicolumn{2}{c}{}  & \multicolumn{1}{c}{ID}  & \multicolumn{5}{c}{\diagnostics (OOD)} & \multicolumn{2}{c}{ID (Val.)} &  \multicolumn{5}{c}{\diagnostics (OOD)}\\
\cmidrule(lr){3-3}  \cmidrule(lr){4-8} \cmidrule(lr){9-10} \cmidrule(lr){11-15}& 
& Test & Lex. & PAS & LS & Kno. & All & Mat. & MisM. & Lex. & PAS & LS & Kno. & All  \\ 
\midrule
& {100\% train}     
& 92.0 &  54.6 & 67.9 & 62.7 & 52.1 & 61.8                  & \textbf{90.2} & \textbf{90.1} & 59.9 & 68.4 & 67.3 & 57.8 & 65.0 \\ 
\midrule[0.03em]
\multirow{3}{*}{\rotatebox{90}{{33\% train}}} & \textsl{random}          
& 91.3 &  53.0 & 66.8 & 59.7 & 50.7 & 60.4                  & 89.8 & 89.2 & 59.3 & 69.6 & 66.5 & 56.3 & 64.6 \\ 
\cmidrule[0.03em]{2-15}
& \noisy  
& 91.8 &  55.2 & \textbf{69.1} & 63.2 & 51.7 & 62.0              & 89.5 & 89.7 & 59.3 & 68.9 & \textbf{69.5} & 58.8 & 65.3 \\
& \challenging  
& \textbf{92.2} &  \textbf{58.5} & 67.9 & \textbf{64.1} &\textbf{54.2} & \textbf{63.5}  & 90.1 & 89.3 & \textbf{63.5} & \textbf{71.0} & 68.9 & \textbf{59.2} & \textbf{66.9} \\
\bottomrule
\end{tabular}
\caption{
ID and OOD accuracies for \roberta-large models trained on different selections of \snli and \mnli; we report the best performance over 3 random seeds (see Appendix \S\ref{app:addition_results} for \snli validation results).
\challenging and \noisy subsets of data promote OOD generalization, at minimal degradation of ID performance.
OOD performance improves across all fine-grained linguistic categories in the \diagnostics set.
}
\label{tab:nli}
\end{table*}
\begin{table}
\small
\centering
\begin{tabular}{llcc}
\toprule
 \multicolumn{2}{c}{}  & \multicolumn{1}{c}{In-dist.}  & \multicolumn{1}{c}{Out-of-dist.}\\
\cmidrule(lr){3-3}  \cmidrule(lr){4-4} & & \qnli Val.  & \adversarial   \\ 
\midrule
& {100\% train}           & 93.7$_{0.3}$ & 81.7$_{0.6}$   \\ 
\midrule[0.03em]
\multirow{3}{*}{\rotatebox{90}{{33\% train}}} & \textsl{random}          & 92.7$_{0.3}$ & 78.3$_{0.4}$ \\
\cmidrule[0.03em]{2-4}
& \noisy            & 93.3$_{0.2}$ & 83.3$_{0.6}$ \\
& \challenging       & 93.8$_{0.3}$ & \textbf{83.9}$_{0.2}$ \\
\bottomrule
\end{tabular}
\caption{
QNLI performance on ID validation and OOD test sets, showing substantial improvements in the latter with a third of the original data.
Reported values are averaged over 3 random seeds, with s.d. as subscripts.
}
\label{tab:qnli}
\end{table}

\paragraph{Results}
\label{sec:results-downstream}

We test our selections on the same datasets from the previous section---\winogrande, \snli, \mnli and \qnli.
We report ID validation performance, and OOD performance on test sets either created independently of the dataset (\diagnostics \cite{wang2019superglue} for \snli and \mnli, and \wsc \cite{levesque2011winograd} for \winogrande), or specifically to be adversarial to the dataset (\adversarial \cite{jia-liang-2017-adversarial} for \qnli); see App.~\S\ref{app:data} for details.

\tabref{winogrande} shows our results on \winogrande.\footnote{The official test set for \winogrande has been filtered with AFLite, making ID evaluation more challenging than OOD. However, we apply all our selection methods (including the AFLite selection) on \winogrande's unfiltered training data.}
Training on the most \challenging data results in the best OOD performance, exceeding that of {100\% train}, even with just a third of the data.
A similar effect is seen with \noisy, as well as its coarse-grained counterpart, low-\correctness.
In each of the three cases, ID performance is also higher than all other 33\% baselines, though we observe some degradation compared to the full training set; this is expected as with larger amounts of data models tend to fit the dataset distribution rather than the task \cite{torralba2011unbiased}.
The only selection methods that underperform the {random} baseline are \textbf{forgetting}, and the ones where we select data the model is highly confident and decisive about ({high-\confidence}, {high-\correctness}, and {low-\variance}).
The latter pattern, as well as our overall results, highlight the important role played by examples which are challenging for the model, i.e., \challenging and \noisy examples.

Given that our selection methods outperform baselines from prior work, we only report {random} and {100\% train} selection baselines on the remaining datasets, where we see similar trends.
\tabref{nli} shows results for \snli and \mnli, where the random selection baseline is already within 1\% of the \textbf{100\% train} baseline.\footnote{The ID performance of all models exceeds human accuracy (88\%) for \snli. However, the difference in ID and OOD performance in SNLI is quite high, showing that there is still room for improvement in the NLI task.}
Selecting 33\% of the most \challenging examples achieves even better ID performance, within 0.2\% of the 100\% train baseline, while exceeding OOD performance substantially on each of the linguistic categories in the \diagnostics test set.\footnote{While \mnli-mismatched is technically out-of-domain, performance is close to matched (ID).}
While \noisy does not perform as well as \challenging on most cases, it still matches or outperforms the 100\% train baseline on OOD test sets.
\tabref{qnli} shows a similar trend for \qnli, where we gain over 2\% performance on the OOD \adversarial test set, with minimal loss in ID accuracy.

\begin{figure*}[t]
     \centering
     \includegraphics[width=\textwidth]{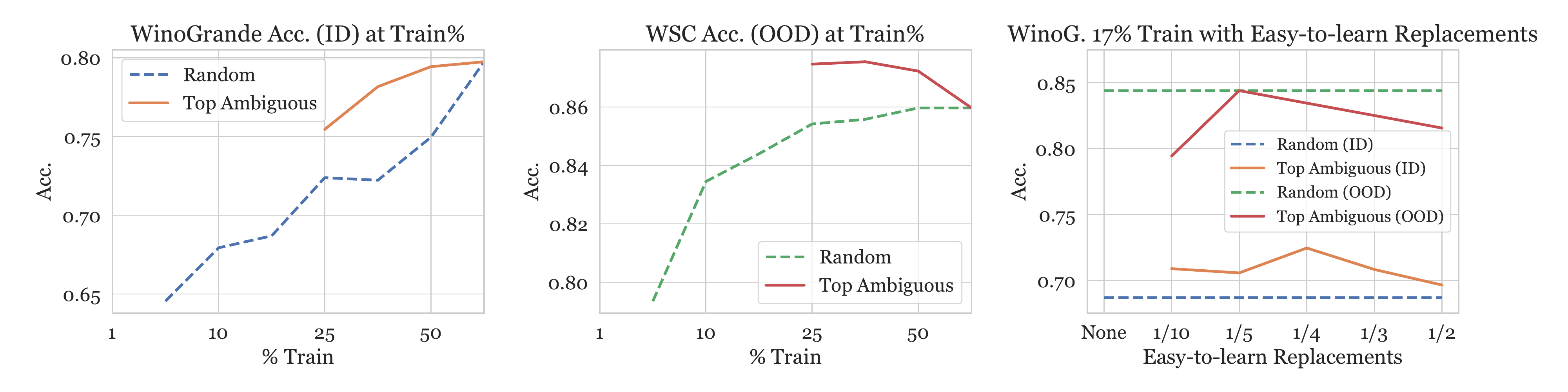}
     \caption{
     ID (left) and OOD (centre) \winogrande performance with increasing \% of \challenging (and randomly-sampled) training data.
     \roberta-large optimization fails when trained on small amounts ($<$ 25\%) of the most \challenging data (results correspond to majority baseline performance and are not shown here, for better visibility).
     (Right) 
     Replacing small amounts of \challenging examples from the 17\% subset with \easy examples results in successful optimization and ID improvements, at the cost of decreased OOD accuracy.
     All reported performances are averaged over 3 random seeds.
     } 
     \label{fig:easy}
 \end{figure*}

Overall, regions revealed by data maps provide ways to substantially improve OOD performance across datasets.
Regional selections of data not only improve model generalization, but also do so using substantially less data, providing a method to potentially speed up training. 
We note, however, that discovering such examples requires computing training dynamics, which involves training a model on the full dataset.
Future directions involve building more efficient data maps, to better fulfill the training speedup potential.

\section{Role of \Easy Instances}
\label{sec:easy}

Data maps uncover \challenging regions, small subsets from which lead to improved OOD performance, with minimal degradation of ID performance (\S\ref{sec:selection}).
We next investigate how performance is affected as we vary the size of the \challenging subsets.
We retrain our model with subsets containing the top 50\%, 33\%, 25\%, 17\%, 10\%, 5\% and 1\% \challenging instances of \winogrande (\figref{easy}, left and center).
Large \challenging subsets (25\% or more) result in high ID and OOD performance.
Surprisingly however, for smaller \challenging subsets (17\% or less), the model performs at chance level, despite random restarts.\footnote{This is common for large models trained on small datasets \cite{Devlin:2019,Phang:2018,Dodge:2020}.}
In contrast, a baseline that randomly selects subsets of similar sizes is able to learn (while naturally performing worse as data decreases, eventually failing at 1\%).
This indicates that \challenging instances alone might be insufficient for learning. 

Given that the model barely struggles with \easy instances (by definition), we next replace some \challenging examples with \easy examples in the 17\% most \challenging subset. 
Interestingly, replacing just a tenth of the \challenging data with \easy instances, the model not only successfully learns, but also outperforms the random selection baseline's ID performance (\figref{easy} right).
This indicates that for successful optimization, it is important to include \textit{easier}-to-learn instances.\todo{AM: This reminds of the literature in adversarial training that demonstrates the benefits of mixing the clean/original and perturbed/adversarial data, but also show that the ratio is important.}
However, with too many replacements, performance starts decreasing again; this trend was seen in the previous section with the high-\confidence baseline (Tab.~\ref{tab:winogrande}).
OOD performance shows a similar trend, but matches or is worse than the baseline.
Selection of the optimal balance of \easy and \challenging examples in low data regimes is an open problem; we defer this exploration to future work.

\section{Detecting Mislabeled Examples}
\label{sec:noise}


Crowdsourced datasets are often subject to noise attributed to incorrect labeled annotations \cite{sheng2008get,krishna2016embracing,ekambaram2017finding}, which may lead to training models that are not representative of the task at hand.
Recent studies have shown that over-parameterized neural networks can fit the incorrect labels blindly \cite{zhang2017rethinking}, which might hurt their generalization ability \cite{hu2020generalization}. 
For large datasets, identifying mislabeled examples can be prohibitively expensive. 
Our data maps provide a semi-automated method to identify such mislabeled instances, without significantly more effort than simply training a model on the data.
We hypothesize that \noisy examples---those with low \confidence---might be mislabeled, as has also been suggested in prior work \cite{manning2011pos,toneva2018empirical}.

\begin{figure}[t]
     \centering
     \includegraphics[width=\columnwidth]{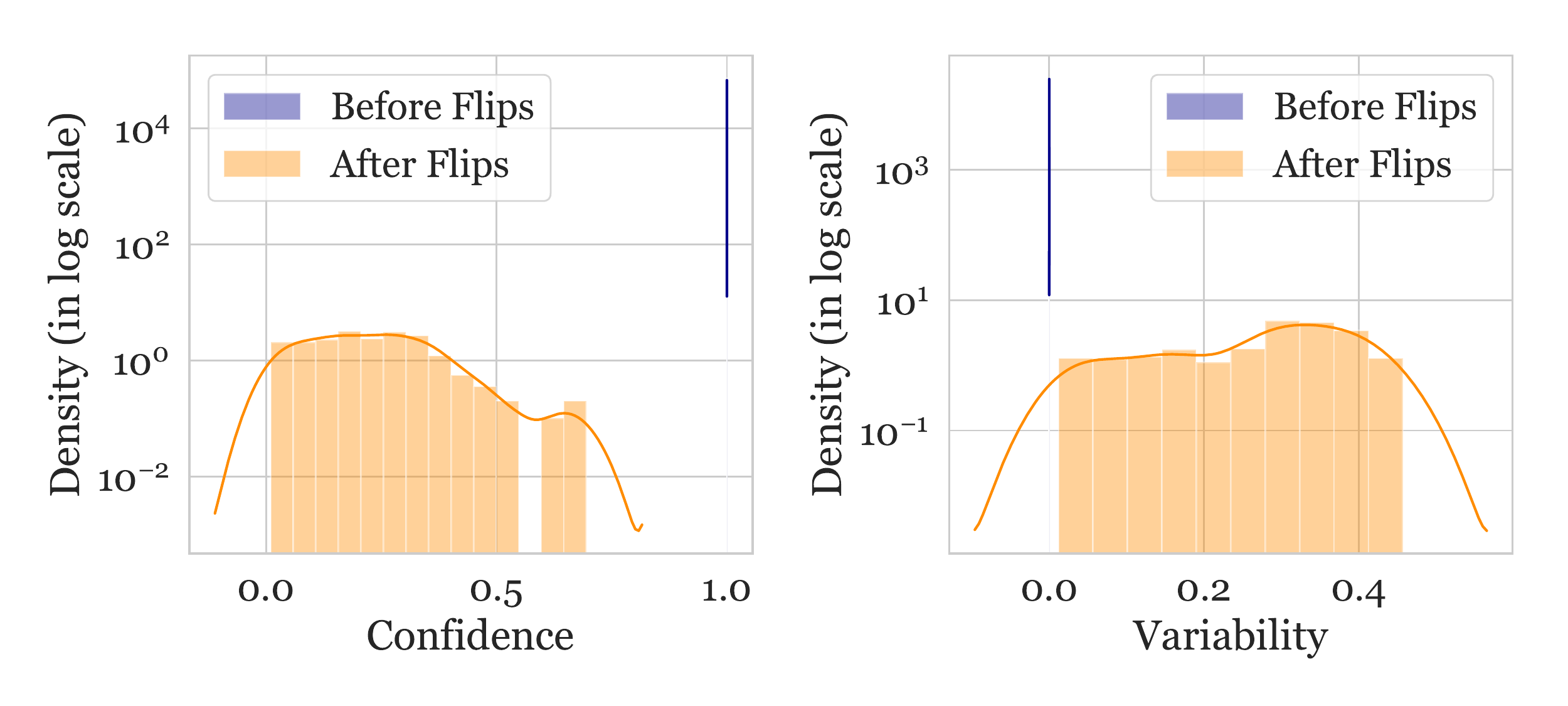}
     \caption{
        Retraining \winogrande with 1\% noised (label-flipped) data changes the training dynamics of the noisy examples. 
        After retraining, there is a noticeable distributional shift towards lower \confidence, with some shift towards higher \variance as well.
        \nascomment{Kind of hard to read / understand. $\Delta$ \confiddence and $\Delta$ \variance densities might be more informative?}
     } 
     \label{fig:flips-wg}
\end{figure}

To verify this hypothesis, we design an experimental setting where we artificially inject noise in the training data, by flipping the labels of 1\% of the training data for \winogrande.
Motivated by our qualitative analysis (Tab~\ref{tab:qualitative}), we select the candidates for flipping from the \easy region---this minimizes the risk of selecting already mislabeled examples.
We retrain \roberta with the partly noised data, and recompute \confidence and \variance of all instances.
\figref{flips-wg} shows the training dynamics measures, before and after re-training. 
Flipped instances move to the lower \confidence regions after retraining, with some movement towards higher \variance.
This indicates that perhaps the \noisy region (low \confidence) of the map contains other mislabeled instances.
We next explore a simple method to automatically detect such instances.

\paragraph{Automatic Noise Detection}
\label{sec:auto_noise}

We train a linear model to classify examples as mislabeled (noise) or not, using a single feature: the \confidence score from the retrained \roberta model on \winogrande.
This model is trained using a balanced training set for this task by sampling equal numbers of noisy (label-flipped) and clean examples from the original train set.
This simple classifier is quite effective---a sanity check evaluation on a similarly constructed test set yields 100\% F1.\footnote{A similar experiment with only \variance scores as features resulted in a much poorer classifier---70\% F1.}

Next, we run the trained noise classifier on the entire original training set, with features extracted from the original training dynamics measures (computed without added noise).
We first observe that despite training on a balanced dataset, our classifier predicts only a few examples as mislabeled---only 31 \winogrande instances (out a total of 40K). 
A similar experiment on \snli results in 15K noisy examples (out of 500K).
These results are encouraging and follow our intuitions that most instances in data are indeed labeled ``correctly''.
Indeed, \winogrande contains a lower portion of noisy examples, as indicated by our data maps (\figref{winogrande}).

We further investigated these trends via a human evaluation on the output of the classifier.
We created an evaluation set by randomly selecting 50 instances from each predicted class as per our classifier. 
Two of the authors re-annotated these 100 instances (without access to the original or predicted labels); some instances were annotated as too ambiguous for the authors. 
After discussions to resolve their differences, both annotators agreed on 96\% of the instances in each dataset.
Using our annotations as a new gold standard, we found that  for \winogrande, 67\% of the instances predicted as noisy by the linear classifier are indeed either mislabeled or ambiguous, compared to only 13\% of the ones predicted as correctly labeled.
Similar patterns are observed for \snli (76\% vs. 4\%). 
\nascomment{provide confusion matrix instead?}
\todo{Table ?? shows some examples identified as noise by the classifier.}

Our results demonstrate the potential of using data maps as a tool to ``clean-up'' datasets, by identifying mislabeled or ambiguous instances.\footnote{In preliminary experiments, retraining \winogrande after removal of noise did not yield a large difference in performance, given the relatively small amount of noise.}
Notably, our results were obtained using a simple method; this encourages exploration of methods that might lead to more accurate noise-detectors.

\section{Training Dynamics as Uncertainty Measures}
\label{sec:analysis}

We introduced data maps, and used training dynamics measures as coordinates for data points in \S\ref{sec:mapping}.
We now take a closer look at these measures, and find intuitive connections with measures of \emph{uncertainty}. 
When a model fails to predict the correct label, the error may come from ambiguity inherent to the example (\emph{intrinsic uncertainty}), but it may also come from the model's limitations (often referred to as \emph{model uncertainty}).\footnote{These are also sometimes called the \emph{aleatoric} and \emph{epistemic} uncertainty, respectively \cite{Gal2016Uncertainty}.}
To understand how examples contribute to a dataset, it is important to separate these two sources of error.

We start by studying the relationship between intrinsic uncertainty and our training dynamics measures.
Human agreement can serve as a proxy for intrinsic uncertainty. 
We estimate human agreement using the multiple human annotations available in \snli's development set.\footnote{Only \snli dev.~and test set have multiple annotations on all instances. We obtain training dynamics with \roberta-large run on train and dev.~combined.}
For each annotator, we compute whether they agree with the majority label from the other four, breaking ties randomly and then averaging over annotators.\footnote{Normally, this provides the minimum-variance unbiased estimate, though \snli's development set throws away examples without a majority, which introduces some bias.}$^,$\footnote{Note that the model only has seen the majority vote, while we take into account all annotator labels to quantify agreement.}

\begin{figure}[t]
    \centering
    \includegraphics[width=\columnwidth]{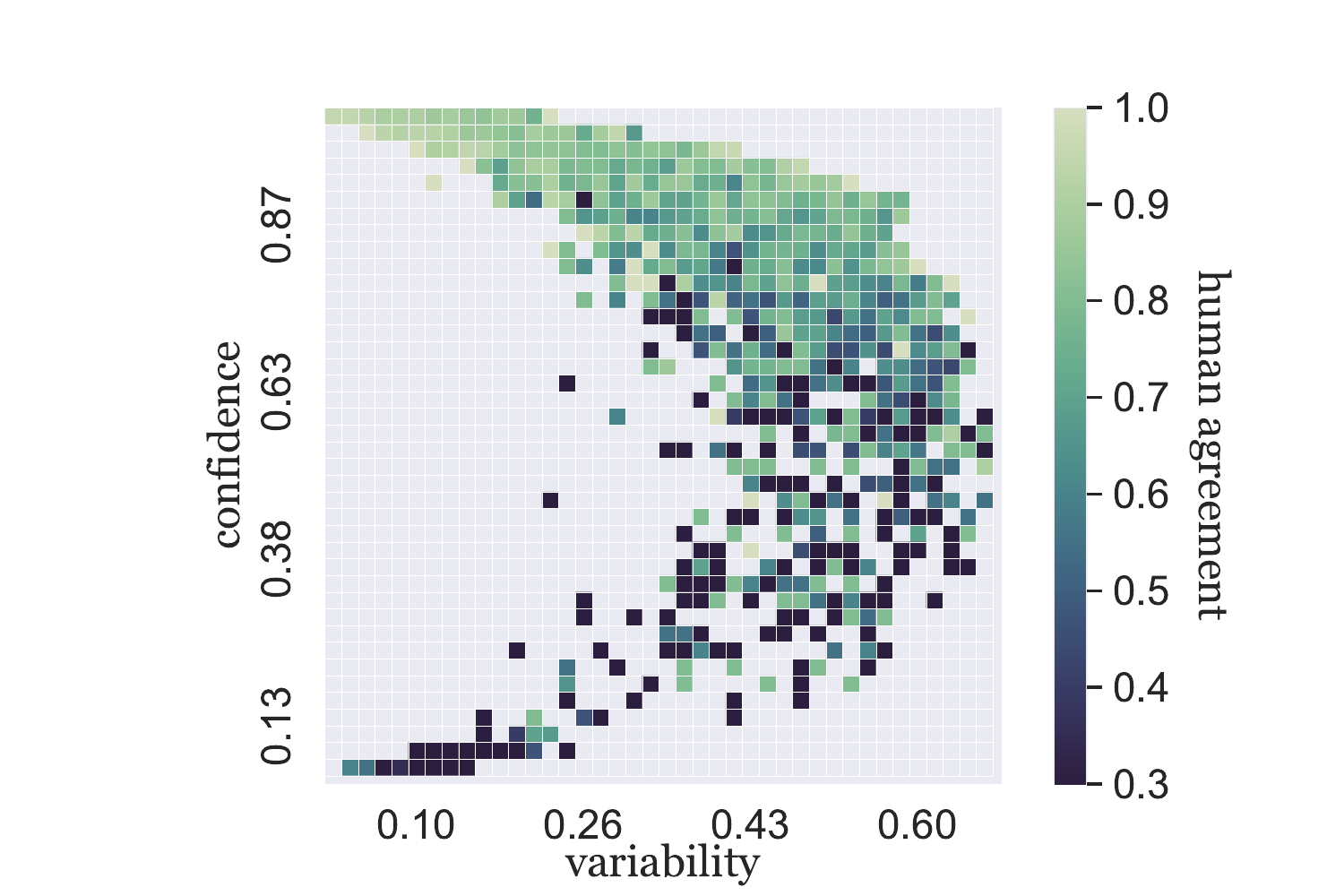}
    \caption{
    Visualizing human agreement on the \snli (dev.~set only) data map reveals its strong relationship to \confidence. 
    Each cell in the heatmap bins examples based on \confidence and \variance, then colors the cell by the mean human agreement.
    } 
    \label{fig:human}
\end{figure}

\figref{human} visualizes the relation between our training dynamics measures (\confidence and \variance) and human agreement, averaged over the examples.
We observe a strong relationship between human agreement and \confidence: high \confidence indicates high agreement between annotators, while low \confidence often indicates disagreement on the example.
In contrast, once \confidence is known, \variance does not provide much information about the agreement.

The connection between our second measure, \variance, and \textit{model uncertainty} is more straightforward: \variance, by definition, captures exactly the uncertainty of the model.
See App.~\ref{app:training-dynamics-vs-dropout} for an additional discussion (with empirical justifications) on connections between training dynamics measures and dropout-based \cite{srivastava2014dropout}, first-principles uncertainty estimates.

These relations are further supported by previous work, which showed that deep ensembles provide well-calibrated uncertainty estimates \cite{lakshminarayanan2017simple, gustafsson2019evaluating, snoek2019can}. 
Generally, such approaches ensemble models trained from scratch; while ensembles of training checkpoints lose some diversity \cite{fort2019deep}, they offer a cheaper alternative capturing some of the benefits \cite{Chen2017CheckpointEE}.
Future work will involve investigation of such alternatives for building data maps.

\section{Related Work}
\label{sec:related}
\todo{Additional relevant references from \citet{toneva2018empirical}, \citet{chang2017active} and  \citet{koh2019accuracy}. AM: If you have extra space, it could be nice to connect to similar (still very different) works that introduce learnable taxonomies, e.g. "Taskonomy: Disentangling Task Transfer Learning" or "TASK2VEC: Task Embedding for Meta-Learning"}

Our work builds data maps using training dynamics measures for scoring data instances.
Loss landscapes \cite{xing2018walk} are similar to training dynamics, but also consider variables from the stochastic optimization algorithm.
\citet{toneva2018empirical} also use training dynamics to find train examples which are frequently ``forgotten'', i.e., misclassified during a later epoch of training, despite being classified correctly earlier; our \correctness metric provides similar discrete scores, and results in models with better performance.
Variants of such approaches address catastrophic forgetting, and are useful for analyzing data instances \cite{pan2020continual,krymolowski2002-distinguishing}.

Prior work has proposed other criteria to score instances.
AFLite \cite{lebras2020adversarial} is an adversarial filtering algorithm which ranks instances based on their ``predictability'', i.e. the ability of simple linear classifiers to predict them correctly. 
While AFLite, among others \cite{Li2019repair,gururangan-etal-2018-annotation}, advocate removing ``easy'' instances from the dataset,  our work shows that \easy instances can be useful.
Similar intuitions have guided other work such as curriculum learning \cite{Bengio:2009} and self-paced learning \cite{kumar2010self,Lee2011LearningTE} where all examples are prioritized based on their ``difficulty''.

Other approaches have used training loss \cite{han2018coteaching,eric2019unsupervisedlabelnoise,shen2019learningbad}, confidence \cite{hovy-etal-2013-learning}, and meta-learning \cite{ren2018learning}, to differentiate instances within datasets.
Perhaps our measures are the closest to those from \citet{chang2017active}; they propose prediction variance and threshold closeness---which correspond to \variance and \confidence, respectively.\footnote{They also consider confidence intervals; our preliminary experiments, with and without, yielded similar results.}
However, they use these measures to reweight all instances, similar to sampling effective batches in online learning \cite{loshchilov2015online}.
Our work, instead, does a hard selection for the purpose of studying different groups within data.

Our methods are also reminiscent of active learning methods \citep{settles2009active,peris-casacuberta2018active,pvs-meyer-2019-data}, such as uncertainty sampling \cite{lewis1994sequential} which selects (unlabeled) data points, which a model trained on a small labeled subset, has least confidence in, or predicts as farthest (in vector space, based on cosine similarity) \citep{sener2018active,wolf2011facility}.
Our approach uses labeled data for selection, similar to core-set selection approaches  \cite{wei-2013-using}.
Active learning approaches could be used in conjunction with data maps to create better datasets, similar to approaches proposed in \citet{mishra2020dqi}.
For instance, creating datasets with more \challenging examples (with respect to a given model) could make it beneficial for OOD generalization.

Data error detection also involves instance scoring.
Influence functions \cite{koh2017understanding}, forgetting events \cite{toneva2018empirical}, cross validation \cite{chen2019understanding}, Shapely values \cite{ghorbani2019data}, and the area-under-margin metric \cite{pleiss2020identifying} have all been used to identify mislabeled examples. 
Some approaches avoid hard examples altogether \cite{Bottou2005-yg,northcutt2017learning} to reduce fit to noisy data.
Our use of training dynamics to locate mislabeled examples involves minimal additional effort beyond training a model on the dataset.

\section{Conclusion}
\label{sec:discussion}

We presented \textit{data maps}: an automatic method to visualize and diagnose large datasets using training dynamics. 
Our data maps for four different datasets reveal similar terrains in each dataset: groups of \challenging instances useful for high performance, \easy instances which aid optimization, and \noisy instances which often correspond to data errors. \todo{consider density based clustering?}
While our maps are based on \roberta-large, the methods to build them are model-agnostic (App.~\S\ref{sec:encoder}).
Our work shows the effectiveness of simple training dynamics measures based on mean and standard deviation; exploration of more sophisticated measures to build data maps is an exciting future direction.
Data maps not only help diagnose and make better use of existing datasets, but also hold potential for guiding the construction of new datasets.
Moreover, data maps could facilitate comparison of different model architectures trained on a given dataset, resulting in alternative evaluation methodologies.
Our implementation is publicly available to facilitate such efforts.\footnote{\url{https://github.com/allenai/cartography}}
\todo{anything about confidence calibration?}

\section*{Acknowledgements}
This research was supported in part by DARPA under the MCS program through NIWC Pacific (N66001-19-2-4031) grant, and by the Allen Distinguished Investigator Award.
We thank the anonymous reviewers, and our colleagues from AI2 and UWNLP, especially Ana Marasovi\'{c}, and Suchin Gururangan, for their helpful feedback.
\bibliography{anthology,emnlp2020}

\begin{thebibliography}{72}
\expandafter\ifx\csname natexlab\endcsname\relax\def\natexlab#1{#1}\fi

\bibitem[{Antol et~al.(2015)Antol, Agrawal, Lu, Mitchell, Batra, Zitnick, and
  Parikh}]{agrawal2015vqa}
Stanislaw Antol, Aishwarya Agrawal, Jiasen Lu, Margaret Mitchell, Dhruv Batra,
  C.~Lawrence Zitnick, and Devi Parikh. 2015.
\newblock \href {https://arxiv.org/abs/1505.00468} {{VQA}: {V}isual {Q}uestion
  {A}nswering}.
\newblock In \emph{ICCV}.

\bibitem[{Arazo et~al.(2019)Arazo, Ortego, Albert, O'Connor, and
  Mcguinness}]{eric2019unsupervisedlabelnoise}
Eric Arazo, Diego Ortego, Paul Albert, Noel O'Connor, and Kevin Mcguinness.
  2019.
\newblock \href {https://arxiv.org/abs/1904.11238} {Unsupervised label noise
  modeling and loss correction}.
\newblock In \emph{ICML}, pages 312--321.

\bibitem[{Bengio et~al.(2009)Bengio, Louradour, Collobert, and
  Weston}]{Bengio:2009}
Yoshua Bengio, J{\'e}r{\^o}me Louradour, Ronan Collobert, and Jason Weston.
  2009.
\newblock \href {https://dl.acm.org/doi/abs/10.1145/1553374.1553380}
  {Curriculum learning}.
\newblock In \emph{ICML}.

\bibitem[{Bottou et~al.(2005)Bottou, Weston, and Bakir}]{Bottou2005-yg}
L{\'e}on Bottou, Jason Weston, and G{\"o}khan~H Bakir. 2005.
\newblock \href
  {https://papers.nips.cc/paper/2695-breaking-svm-complexity-with-cross-training}
  {Breaking {SVM} complexity with {Cross-Training}}.
\newblock In \emph{NeurIPS}, pages 81--88. MIT Press.

\bibitem[{Bowman et~al.(2015)Bowman, Angeli, Potts, and
  Manning}]{bowman-etal-2015-large}
Samuel~R. Bowman, Gabor Angeli, Christopher Potts, and Christopher~D. Manning.
  2015.
\newblock \href {https://doi.org/10.18653/v1/D15-1075} {A large annotated
  corpus for learning natural language inference}.
\newblock In \emph{Proceedings of the 2015 Conference on Empirical Methods in
  Natural Language Processing}, pages 632--642, Lisbon, Portugal. Association
  for Computational Linguistics.

\bibitem[{Chang et~al.(2017)Chang, Learned-Miller, and
  McCallum}]{chang2017active}
Haw-Shiuan Chang, Erik Learned-Miller, and Andrew McCallum. 2017.
\newblock \href
  {http://papers.nips.cc/paper/6701-active-bias-training-more-accurate-neural-networks-by-emphasizing-high-variance-samples}
  {Active bias: Training more accurate neural networks by emphasizing high
  variance samples}.
\newblock In \emph{NeurIPS}, pages 1002--1012.

\bibitem[{Chen et~al.(2017{\natexlab{a}})Chen, Lundberg, and
  Lee}]{Chen2017CheckpointEE}
Hugh Chen, Scott Lundberg, and Su-In Lee. 2017{\natexlab{a}}.
\newblock \href {https://arxiv.org/abs/1710.03282} {Checkpoint ensembles:
  Ensemble methods from a single training process}.
\newblock ArXiv:1710.03282.

\bibitem[{Chen et~al.(2019)Chen, Liao, Chen, and Zhang}]{chen2019understanding}
Pengfei Chen, Benben Liao, Guangyong Chen, and Shengyu Zhang. 2019.
\newblock \href {https://arxiv.org/abs/1905.05040} {Understanding and utilizing
  deep neural networks trained with noisy labels}.
\newblock In \emph{{ICML}}, volume~97 of \emph{Proceedings of Machine Learning
  Research}, pages 1062--1070. {PMLR}.

\bibitem[{Chen et~al.(2017{\natexlab{b}})Chen, Zhu, Ling, Wei, Jiang, and
  Inkpen}]{chen-etal-2017-enhanced}
Qian Chen, Xiaodan Zhu, Zhen-Hua Ling, Si~Wei, Hui Jiang, and Diana Inkpen.
  2017{\natexlab{b}}.
\newblock \href {https://doi.org/10.18653/v1/P17-1152} {Enhanced {LSTM} for
  natural language inference}.
\newblock In \emph{Proceedings of the 55th Annual Meeting of the Association
  for Computational Linguistics (Volume 1: Long Papers)}, pages 1657--1668,
  Vancouver, Canada. Association for Computational Linguistics.

\bibitem[{Devlin et~al.(2019)Devlin, Chang, Lee, and Toutanova}]{Devlin:2019}
Jacob Devlin, Ming-Wei Chang, Kenton Lee, and Kristina Toutanova. 2019.
\newblock \href {https://doi.org/10.18653/v1/N19-1423} {{BERT}: Pre-training of
  deep bidirectional transformers for language understanding}.
\newblock In \emph{NAACL}.

\bibitem[{Dodge et~al.(2019)Dodge, Gururangan, Card, Schwartz, and
  Smith}]{dodge2019show}
Jesse Dodge, Suchin Gururangan, Dallas Card, Roy Schwartz, and Noah~A. Smith.
  2019.
\newblock \href {https://arxiv.org/abs/1909.03004} {Show your work: Improved
  reporting of experimental results}.
\newblock In \emph{EMNLP}.

\bibitem[{Dodge et~al.(2020)Dodge, Ilharco, Schwartz, Farhadi, Hajishirzi, and
  Smith}]{Dodge:2020}
Jesse Dodge, Gabriel Ilharco, Roy Schwartz, Ali Farhadi, Hannaneh Hajishirzi,
  and Noah~A. Smith. 2020.
\newblock \href {https://arxiv.org/abs/2002.06305} {Fine-tuning pretrained
  language models: Weight initializations, data orders, and early stopping}.
\newblock {arXiv}:2002.06305.

\bibitem[{{Ekambaram} et~al.(2017){Ekambaram}, {Goldgof}, and
  {Hall}}]{ekambaram2017finding}
R.~{Ekambaram}, D.~B. {Goldgof}, and L.~O. {Hall}. 2017.
\newblock \href {https://ieeexplore.ieee.org/document/8122985/} {Finding label
  noise examples in large scale datasets}.
\newblock In \emph{SMC}, pages 2420--2424.

\bibitem[{Fort et~al.(2019)Fort, Hu, and Lakshminarayanan}]{fort2019deep}
Stanislav Fort, Huiyi Hu, and Balaji Lakshminarayanan. 2019.
\newblock \href {https://arxiv.org/abs/1912.02757} {Deep ensembles: A loss
  landscape perspective}.
\newblock ArXiv preprint arXiv:1912.02757.

\bibitem[{Gal(2016)}]{Gal2016Uncertainty}
Yarin Gal. 2016.
\newblock \href {http://mlg.eng.cam.ac.uk/yarin/thesis/thesis.pdf}
  {\emph{Uncertainty in Deep Learning}}.
\newblock Ph.D. thesis, University of Cambridge.

\bibitem[{Gal and Ghahramani(2016)}]{gal2016dropout}
Yarin Gal and Zoubin Ghahramani. 2016.
\newblock \href {http://proceedings.mlr.press/v48/gal16.html} {Dropout as a
  bayesian approximation: Representing model uncertainty in deep learning}.
\newblock In \emph{ICML}, volume~48, pages 1050--1059. PMLR.

\bibitem[{Ghorbani and Zou(2019)}]{ghorbani2019data}
Amirata Ghorbani and James Zou. 2019.
\newblock \href {http://arxiv.org/abs/1904.02868} {Data shapley: Equitable
  valuation of data for machine learning}.

\bibitem[{Gururangan et~al.(2018)Gururangan, Swayamdipta, Levy, Schwartz,
  Bowman, and Smith}]{gururangan-etal-2018-annotation}
Suchin Gururangan, Swabha Swayamdipta, Omer Levy, Roy Schwartz, Samuel Bowman,
  and Noah~A. Smith. 2018.
\newblock \href {https://doi.org/10.18653/v1/N18-2017} {Annotation artifacts in
  natural language inference data}.
\newblock In \emph{Proceedings of the 2018 Conference of the North {A}merican
  Chapter of the Association for Computational Linguistics: Human Language
  Technologies, Volume 2 (Short Papers)}, pages 107--112, New Orleans,
  Louisiana. Association for Computational Linguistics.

\bibitem[{Gustafsson et~al.(2019)Gustafsson, Danelljan, and
  Sch{\"o}n}]{gustafsson2019evaluating}
Fredrik~K Gustafsson, Martin Danelljan, and Thomas~B Sch{\"o}n. 2019.
\newblock \href {https://arxiv.org/abs/1906.01620} {Evaluating scalable
  bayesian deep learning methods for robust computer vision}.
\newblock ArXiv preprint arXiv:1906.01620.

\bibitem[{Han et~al.(2018)Han, Yao, Yu, Niu, Xu, Hu, Tsang, and
  Sugiyama}]{han2018coteaching}
Bo~Han, Quanming Yao, Xingrui Yu, Gang Niu, Miao Xu, Weihua Hu, Ivor~W. Tsang,
  and Masashi Sugiyama. 2018.
\newblock \href {https://arxiv.org/abs/1804.06872} {Co-teaching: Robust
  training of deep neural networks with extremely noisy labels}.
\newblock In \emph{NeurIPS}, pages 8536--8546.

\bibitem[{Hendrycks et~al.(2020)Hendrycks, Liu, Wallace, Dziedzic, Krishnan,
  and Song}]{hendrycks2020pretrained}
Dan Hendrycks, Xiaoyuan Liu, Eric Wallace, Adam Dziedzic, Rishabh Krishnan, and
  Dawn Song. 2020.
\newblock \href {https://arxiv.org/abs/2004.06100} {Pretrained transformers
  improve out-of-distribution robustness}.
\newblock ArXiv preprint arXiv:2004.06100.

\bibitem[{Hovy et~al.(2013)Hovy, Berg-Kirkpatrick, Vaswani, and
  Hovy}]{hovy-etal-2013-learning}
Dirk Hovy, Taylor Berg-Kirkpatrick, Ashish Vaswani, and Eduard Hovy. 2013.
\newblock \href {https://www.aclweb.org/anthology/N13-1132} {Learning whom to
  trust with {MACE}}.
\newblock In \emph{Proceedings of the 2013 Conference of the North {A}merican
  Chapter of the Association for Computational Linguistics: Human Language
  Technologies}, pages 1120--1130, Atlanta, Georgia. Association for
  Computational Linguistics.

\bibitem[{Hu et~al.(2020)Hu, Li, and Yu}]{hu2020generalization}
Wei Hu, Zhiyuan Li, and Dingli Yu. 2020.
\newblock \href {https://arxiv.org/abs/1905.11368} {Simple and effective
  regularization methods for training on noisily labeled data with
  generalization guarantee}.
\newblock In \emph{{ICLR}}. OpenReview.net.

\bibitem[{Jia and Liang(2017)}]{jia-liang-2017-adversarial}
Robin Jia and Percy Liang. 2017.
\newblock \href {https://doi.org/10.18653/v1/D17-1215} {Adversarial examples
  for evaluating reading comprehension systems}.
\newblock In \emph{Proceedings of the 2017 Conference on Empirical Methods in
  Natural Language Processing}, pages 2021--2031, Copenhagen, Denmark.
  Association for Computational Linguistics.

\bibitem[{Joshi et~al.(2009)Joshi, Porikli, and
  Papanikolopoulos}]{joshi-2009-multi}
Ajay~J Joshi, Fatih Porikli, and Nikolaos Papanikolopoulos. 2009.
\newblock \href {https://ieeexplore.ieee.org/abstract/document/5206627}
  {Multi-class active learning for image classification}.
\newblock In \emph{CVPR}, pages 2372--2379. IEEE.

\bibitem[{Kaplan et~al.(2020)Kaplan, McCandlish, Henighan, Brown, Chess, Child,
  Gray, Radford, Wu, and Amodei}]{kaplan2020scaling}
Jared Kaplan, Sam McCandlish, Tom Henighan, Tom~B. Brown, Benjamin Chess, Rewon
  Child, Scott Gray, Alec Radford, Jeffrey Wu, and Dario Amodei. 2020.
\newblock \href {http://arxiv.org/abs/2001.08361} {Scaling laws for neural
  language models}.

\bibitem[{Kingma and Ba(2014)}]{kingma2014adam}
Diederik~P. Kingma and Jimmy Ba. 2014.
\newblock \href {https://arxiv.org/abs/1412.6980} {Adam: A method for
  stochastic optimization}.
\newblock ArXiv:1412.6980.

\bibitem[{Koh and Liang(2017)}]{koh2017understanding}
Pang~Wei Koh and Percy Liang. 2017.
\newblock \href {https://dl.acm.org/doi/10.5555/3305381.3305576} {Understanding
  black-box predictions via influence functions}.
\newblock In \emph{ICML}, pages 1885--1894. JMLR. org.

\bibitem[{Krishna et~al.(2016)Krishna, Hata, Chen, Kravitz, Shamma, Li, and
  Bernstein}]{krishna2016embracing}
Ranjay~A. Krishna, Kenji Hata, Stephanie Chen, Joshua Kravitz, David~A. Shamma,
  Fei{-}Fei Li, and Michael~S. Bernstein. 2016.
\newblock \href {https://arxiv.org/pdf/1602.04506} {Embracing error to enable
  rapid crowdsourcing}.
\newblock In \emph{CHI}, pages 3167--3179. {ACM}.

\bibitem[{Krymolowski(2002)}]{krymolowski2002-distinguishing}
Yuval Krymolowski. 2002.
\newblock \href {https://www.aclweb.org/anthology/W02-2015} {Distinguishing
  easy and hard instances}.
\newblock In \emph{{COLING}}.

\bibitem[{Kumar et~al.(2010)Kumar, Packer, and Koller}]{kumar2010self}
M~Pawan Kumar, Benjamin Packer, and Daphne Koller. 2010.
\newblock \href
  {https://papers.nips.cc/paper/3923-self-paced-learning-for-latent-variable-models}
  {Self-paced learning for latent variable models}.
\newblock In \emph{NeurIPS}, pages 1189--1197.

\bibitem[{Lakshminarayanan et~al.(2017)Lakshminarayanan, Pritzel, and
  Blundell}]{lakshminarayanan2017simple}
Balaji Lakshminarayanan, Alexander Pritzel, and Charles Blundell. 2017.
\newblock \href
  {http://papers.nips.cc/paper/7219-simple-and-scalable-predictive-uncertainty-estimation-using-deep-ensembles}
  {Simple and scalable predictive uncertainty estimation using deep ensembles}.
\newblock In \emph{NeurIPS}, pages 6402--6413.

\bibitem[{LeBras et~al.(2020)LeBras, Swayamdipta, Bhagavatula, Zellers, Peters,
  Sabharwal, and Choi}]{lebras2020adversarial}
Ronan LeBras, Swabha Swayamdipta, Chandra Bhagavatula, Rowan Zellers,
  Matthew~E. Peters, Ashish Sabharwal, and Yejin Choi. 2020.
\newblock \href {https://arxiv.org/abs/2002.04108} {Adversarial filters of
  dataset biases}.
\newblock In \emph{ICML}.

\bibitem[{Lee and Grauman(2011)}]{Lee2011LearningTE}
Yong~Jae Lee and Kristen Grauman. 2011.
\newblock \href {https://ieeexplore.ieee.org/document/5995523} {Learning the
  easy things first: Self-paced visual category discovery}.
\newblock \emph{CVPR}, pages 1721--1728.

\bibitem[{Levesque et~al.(2011)Levesque, Davis, and
  Morgenstern}]{levesque2011winograd}
Hector~J Levesque, Ernest Davis, and Leora Morgenstern. 2011.
\newblock \href
  {https://www.aaai.org/ocs/index.php/KR/KR12/paper/viewPaper/4492} {The
  {W}inograd schema challenge}.
\newblock In \emph{{AAAI}}, volume~46, page~47.

\bibitem[{Lewis and Gale(1994)}]{lewis1994sequential}
David~D. Lewis and William~A. Gale. 1994.
\newblock \href {https://arxiv.org/abs/cmp-lg/9407020} {A sequential algorithm
  for training text classifiers}.
\newblock In \emph{SIGIR}, SIGIR ’94, page 3–12, Berlin, Heidelberg.
  Springer-Verlag.

\bibitem[{Li and Vasconcelos(2019)}]{Li2019repair}
Yi~Li and Nuno Vasconcelos. 2019.
\newblock \href {https://doi.org/10.1109/cvpr.2019.00980} {{REPAIR}: Removing
  representation bias by dataset resampling}.
\newblock IEEE.

\bibitem[{Linzen(2020)}]{linzen2020can}
Tal Linzen. 2020.
\newblock \href {https://arxiv.org/abs/2005.00955} {How can we accelerate
  progress towards human-like linguistic generalization?}
\newblock In \emph{ACL}.

\bibitem[{Liu et~al.(2019)Liu, Ott, Goyal, Du, Joshi, Chen, Levy, Lewis,
  Zettlemoyer, and Stoyanov}]{Liu2019RoBERTaAR}
Yinhan Liu, Myle Ott, Naman Goyal, Jingfei Du, Mandar~S. Joshi, Danqi Chen,
  Omer Levy, Mike Lewis, Luke~S. Zettlemoyer, and Veselin Stoyanov. 2019.
\newblock \href {https://arxiv.org/abs/1907.11692} {{RoBERTa}: {A} robustly
  optimized {BERT} pretraining approach}.
\newblock ArXiv:1907.11692.

\bibitem[{Loshchilov and Hutter(2016)}]{loshchilov2015online}
Ilya Loshchilov and Frank Hutter. 2016.
\newblock \href {https://arxiv.org/abs/1511.06343} {Online batch selection for
  faster training of neural networks}.
\newblock In \emph{ICLR}.

\bibitem[{Manning(2011)}]{manning2011pos}
Christopher~D. Manning. 2011.
\newblock \href {https://dl.acm.org/doi/10.5555/1964799.1964816}
  {Part-of-speech tagging from 97\% to 100\%: Is it time for some linguistics?}
\newblock In \emph{CICLing}, CICLing’11, page 171–189, Berlin, Heidelberg.
  Springer-Verlag.

\bibitem[{Mishra et~al.(2020)Mishra, Arunkumar, Sachdeva, Bryan, and
  Baral}]{mishra2020dqi}
Swaroop Mishra, Anjana Arunkumar, Bhavdeep Sachdeva, Chris Bryan, and Chitta
  Baral. 2020.
\newblock \href {https://arxiv.org/abs/2005.00816} {{DQI: Measuring Data
  Quality in NLP}}.

\bibitem[{Northcutt et~al.(2017)Northcutt, Wu, and
  Chuang}]{northcutt2017learning}
Curtis~G. Northcutt, Tailin Wu, and Isaac~L. Chuang. 2017.
\newblock \href {http://arxiv.org/abs/1705.01936} {Learning with confident
  examples: Rank pruning for robust classification with noisy labels}.

\bibitem[{Pan et~al.(2020)Pan, Swaroop, Immer, Eschenhagen, Turner, and
  Khan}]{pan2020continual}
Pingbo Pan, Siddharth Swaroop, Alexander Immer, Runa Eschenhagen, Richard~E.
  Turner, and Mohammad~Emtiyaz Khan. 2020.
\newblock \href {https://arxiv.org/abs/2004.14070} {Continual deep learning by
  functional regularisation of memorable past}.

\bibitem[{Peris and Casacuberta(2018)}]{peris-casacuberta2018active}
{\'A}lvaro Peris and Francisco Casacuberta. 2018.
\newblock \href {https://doi.org/10.18653/v1/K18-1015} {Active learning for
  interactive neural machine translation of data streams}.
\newblock In \emph{CoNLL}, pages 151--160.

\bibitem[{Phang et~al.(2018)Phang, F{\'e}vry, and Bowman}]{Phang:2018}
Jason Phang, Thibault F{\'e}vry, and Samuel~R. Bowman. 2018.
\newblock \href {https://arxiv.org/abs/1811.01088} {Sentence encoders on
  {STILTs}: Supplementary training on intermediate labeled-data tasks}.
\newblock {arXiv}:1811.01088.

\bibitem[{Pleiss et~al.(2020)Pleiss, Zhang, Elenberg, and
  Weinberger}]{pleiss2020identifying}
Geoff Pleiss, Tianyi Zhang, Ethan~R. Elenberg, and Kilian~Q. Weinberger. 2020.
\newblock \href {https://arxiv.org/abs/2001.10528} {Identifying mislabeled data
  using the area under the margin ranking}.

\bibitem[{P.V.S and Meyer(2019)}]{pvs-meyer-2019-data}
Avinesh P.V.S and Christian~M. Meyer. 2019.
\newblock \href {https://doi.org/10.18653/v1/N19-1262} {Data-efficient neural
  text compression with interactive learning}.
\newblock In \emph{NAACL}, pages 2543--2554. Association for Computational
  Linguistics.

\bibitem[{Rajpurkar et~al.(2016)Rajpurkar, Zhang, Lopyrev, and
  Liang}]{Rajpurkar2016SQuAD10}
Pranav Rajpurkar, Jian Zhang, Konstantin Lopyrev, and Percy Liang. 2016.
\newblock \href {https://www.aclweb.org/anthology/D16-1264} {{SQuAD:} 100, 000+
  questions for machine comprehension of text}.
\newblock In \emph{EMNLP}.

\bibitem[{Ren et~al.(2018)Ren, Zeng, Yang, and Urtasun}]{ren2018learning}
Mengye Ren, Wenyuan Zeng, Bin Yang, and Raquel Urtasun. 2018.
\newblock \href {http://arxiv.org/abs/1803.09050} {Learning to reweight
  examples for robust deep learning}.

\bibitem[{Russakovsky et~al.(2015)Russakovsky, Deng, Su, Krause, Satheesh, Ma,
  Huang, Karpathy, Khosla, Bernstein, Berg, and
  Fei-Fei}]{Russakovsky2015Imagenet}
Olga Russakovsky, Jia Deng, Hao Su, Jonathan Krause, Sanjeev Satheesh, Sean Ma,
  Zhiheng Huang, Andrej Karpathy, Aditya Khosla, Michael Bernstein,
  Alexander~C. Berg, and Li~Fei-Fei. 2015.
\newblock \href {https://doi.org/10.1007/s11263-015-0816-y} {{ImageNet Large
  Scale Visual Recognition Challenge}}.
\newblock \emph{IJCV}.

\bibitem[{Sakaguchi et~al.(2020)Sakaguchi, Bras, Bhagavatula, and
  Choi}]{sakaguchi2019winogrande}
Keisuke Sakaguchi, Ronan~Le Bras, Chandra Bhagavatula, and Yejin Choi. 2020.
\newblock \href {https://arxiv.org/abs/1907.10641} {Winogrande: An adversarial
  winograd schema challenge at scale}.
\newblock In \emph{AAAI}.

\bibitem[{Sener and Savarese(2018)}]{sener2018active}
Ozan Sener and Silvio Savarese. 2018.
\newblock \href {https://openreview.net/forum?id=H1aIuk-RW} {Active learning
  for convolutional neural networks: A core-set approach}.
\newblock In \emph{ICLR}.

\bibitem[{Settles(2009)}]{settles2009active}
Burr Settles. 2009.
\newblock \href {http://burrsettles.com/pub/settles.activelearning.pdf} {Active
  learning literature survey}.
\newblock Technical report, University of Wisconsin-Madison Department of
  Computer Sciences.

\bibitem[{Shen and Sanghavi(2019)}]{shen2019learningbad}
Yanyao Shen and Sujay Sanghavi. 2019.
\newblock \href {https://arxiv.org/abs/1810.11874} {Learning with bad training
  data via iterative trimmed loss minimization}.
\newblock In \emph{ICML}, pages 5739--5748.

\bibitem[{Sheng et~al.(2008)Sheng, Provost, and Ipeirotis}]{sheng2008get}
Victor~S Sheng, Foster Provost, and Panagiotis~G Ipeirotis. 2008.
\newblock \href {https://dl.acm.org/doi/abs/10.1145/1401890.1401965} {Get
  another label? improving data quality and data mining using multiple, noisy
  labelers}.
\newblock In \emph{SIGKDD}, pages 614--622.

\bibitem[{Shrivastava et~al.(2016)Shrivastava, Gupta, and
  Girshick}]{shrivastava2016training}
Abhinav Shrivastava, Abhinav Gupta, and Ross Girshick. 2016.
\newblock \href
  {https://www.cv-foundation.org/openaccess/content_cvpr_2016/html/Shrivastava_Training_Region-Based_Object_CVPR_2016_paper.html}
  {Training region-based object detectors with online hard example mining}.
\newblock In \emph{CVPR}, pages 761--769.

\bibitem[{Snoek et~al.(2019)Snoek, Ovadia, Fertig, Lakshminarayanan, Nowozin,
  Sculley, Dillon, Ren, and Nado}]{snoek2019can}
Jasper Snoek, Yaniv Ovadia, Emily Fertig, Balaji Lakshminarayanan, Sebastian
  Nowozin, D~Sculley, Joshua Dillon, Jie Ren, and Zachary Nado. 2019.
\newblock \href
  {http://papers.nips.cc/paper/9547-can-you-trust-your-models-uncertainty-evaluating-predictive-uncertainty-under-dataset-shift}
  {Can you trust your model's uncertainty? evaluating predictive uncertainty
  under dataset shift}.
\newblock In \emph{NeurIPS}, pages 13969--13980.

\bibitem[{Srivastava et~al.(2014)Srivastava, Hinton, Krizhevsky, Sutskever, and
  Salakhutdinov}]{srivastava2014dropout}
Nitish Srivastava, Geoffrey Hinton, Alex Krizhevsky, Ilya Sutskever, and Ruslan
  Salakhutdinov. 2014.
\newblock \href {https://jmlr.org/papers/v15/srivastava14a.html} {Dropout: a
  simple way to prevent neural networks from overfitting}.
\newblock \emph{The journal of machine learning research}, 15(1):1929--1958.

\bibitem[{Talmor et~al.(2019)Talmor, Elazar, Goldberg, and
  Berant}]{Talmor2019oLMpicsO}
Alon Talmor, Yanai Elazar, Yoav Goldberg, and Jonathan Berant. 2019.
\newblock \href {https://arxiv.org/abs/1912.13283} {olmpics - on what language
  model pre-training captures}.
\newblock ArXiv:1912.13283.

\bibitem[{Toneva et~al.(2018)Toneva, Sordoni, des Combes, Trischler, Bengio,
  and Gordon}]{toneva2018empirical}
Mariya Toneva, Alessandro Sordoni, Remi~Tachet des Combes, Adam Trischler,
  Yoshua Bengio, and Geoffrey~J Gordon. 2018.
\newblock \href {https://openreview.net/forum?id=BJlxm30cKm} {An empirical
  study of example forgetting during deep neural network learning}.
\newblock In \emph{ICLR}.

\bibitem[{Torralba and Efros(2011)}]{torralba2011unbiased}
Antonio Torralba and Alexei~A Efros. 2011.
\newblock \href {https://ieeexplore.ieee.org/abstract/document/5995347}
  {Unbiased look at dataset bias}.
\newblock In \emph{CVPR 2011}, pages 1521--1528. IEEE.

\bibitem[{Vodrahalli et~al.(2018)Vodrahalli, Li, and
  Malik}]{vodrahalli2018training}
Kailas Vodrahalli, Ke~Li, and Jitendra Malik. 2018.
\newblock \href {https://arxiv.org/abs/1811.12569} {Are all training examples
  created equal? an empirical study}.
\newblock ArXiv:1811.12569.

\bibitem[{Wang et~al.(2019)Wang, Pruksachatkun, Nangia, Singh, Michael, Hill,
  Levy, and Bowman}]{wang2019superglue}
Alex Wang, Yada Pruksachatkun, Nikita Nangia, Amanpreet Singh, Julian Michael,
  Felix Hill, Omer Levy, and Samuel~R. Bowman. 2019.
\newblock \href {https://arxiv.org/abs/1905.00537} {Super{GLUE}: A stickier
  benchmark for general-purpose language understanding systems}.
\newblock In \emph{NeurIPS}.

\bibitem[{Wang et~al.(2018)Wang, Singh, Michael, Hill, Levy, and
  Bowman}]{wang-etal-2018-glue}
Alex Wang, Amanpreet Singh, Julian Michael, Felix Hill, Omer Levy, and Samuel
  Bowman. 2018.
\newblock \href {https://doi.org/10.18653/v1/W18-5446} {{GLUE}: A multi-task
  benchmark and analysis platform for natural language understanding}.
\newblock In \emph{Proceedings of the 2018 {EMNLP} Workshop {B}lackbox{NLP}:
  Analyzing and Interpreting Neural Networks for {NLP}}, pages 353--355,
  Brussels, Belgium. Association for Computational Linguistics.

\bibitem[{Wei et~al.(2013)Wei, Liu, Kirchhoff, and Bilmes}]{wei-2013-using}
Kai Wei, Yuzong Liu, Katrin Kirchhoff, and Jeff Bilmes. 2013.
\newblock \href {https://www.aclweb.org/anthology/N13-1086/} {Using document
  summarization techniques for speech data subset selection}.
\newblock In \emph{Proceedings of the 2013 Conference of the North American
  Chapter of the Association for Computational Linguistics: Human Language
  Technologies}, pages 721--726.

\bibitem[{Williams et~al.(2018)Williams, Nangia, and
  Bowman}]{williams-etal-2018-broad}
Adina Williams, Nikita Nangia, and Samuel Bowman. 2018.
\newblock \href {https://doi.org/10.18653/v1/N18-1101} {A broad-coverage
  challenge corpus for sentence understanding through inference}.
\newblock In \emph{Proceedings of the 2018 Conference of the North {A}merican
  Chapter of the Association for Computational Linguistics: Human Language
  Technologies, Volume 1 (Long Papers)}, pages 1112--1122, New Orleans,
  Louisiana. Association for Computational Linguistics.

\bibitem[{Wolf(2011)}]{wolf2011facility}
Gert~W Wolf. 2011.
\newblock \href
  {https://www.tandfonline.com/doi/abs/10.1080/13658816.2010.528422?journalCode=tgis20}
  {Facility location: concepts, models, algorithms and case studies. series:
  Contributions to management science}.
\newblock \emph{IJGIS}, 25(2):331--333.

\bibitem[{Wolf et~al.(2019)Wolf, Debut, Sanh, Chaumond, Delangue, Moi, Cistac,
  Rault, Louf, Funtowicz, and Brew}]{Wolf2019HuggingFacesTS}
Thomas Wolf, Lysandre Debut, Victor Sanh, Julien Chaumond, Clement Delangue,
  Anthony Moi, Pierric Cistac, Tim Rault, R'emi Louf, Morgan Funtowicz, and
  Jamie Brew. 2019.
\newblock \href {https://arxiv.org/abs/1910.03771} {Huggingface's transformers:
  State-of-the-art natural language processing}.
\newblock ArXiv:1910.03771.

\bibitem[{Xing et~al.(2018)Xing, Arpit, Tsirigotis, and Bengio}]{xing2018walk}
Chen Xing, Devansh Arpit, Christos Tsirigotis, and Yoshua Bengio. 2018.
\newblock \href {https://arxiv.org/abs/1802.08770} {A walk with {SGD}}.

\bibitem[{Yogatama et~al.(2019)Yogatama, d'Autume, Connor, Kocisky,
  Chrzanowski, Kong, Lazaridou, Ling, Yu, Dyer et~al.}]{yogatama2019learning}
Dani Yogatama, Cyprien de~Masson d'Autume, Jerome Connor, Tomas Kocisky, Mike
  Chrzanowski, Lingpeng Kong, Angeliki Lazaridou, Wang Ling, Lei Yu, Chris
  Dyer, et~al. 2019.
\newblock \href {https://arxiv.org/abs/2005.00955} {Learning and evaluating
  general linguistic intelligence}.
\newblock ArXiv preprint arXiv:1901.11373.

\bibitem[{Zhang et~al.(2017)Zhang, Bengio, Hardt, Recht, and
  Vinyals}]{zhang2017rethinking}
Chiyuan Zhang, Samy Bengio, Moritz Hardt, Benjamin Recht, and Oriol Vinyals.
  2017.
\newblock \href {https://arxiv.org/abs/1611.03530} {Understanding deep learning
  requires rethinking generalization}.
\newblock In \emph{ICLR}. OpenReview.net.

\end{thebibliography}
\bibliographystyle{acl_natbib}

\clearpage
\appendix
\section{Supplemental Material}
\label{sec:supplemental}

\subsection{Training Dynamics Computation}
\label{sec:efficient_training_dynamics}
Both \confidence and \variance are computed across epochs, but could alternatively be computed over other granularities,  e.g.~over every few steps of optimization.
This might enable more efficient computation of the same.
However, care must be taken to ignore the first few steps till optimization stabilizes.
In our experiments, we considered all epochs including the first to compute the training dynamics, since the first epoch contains multiple steps of optimization for large training sets.

\begin{figure}[t]
     \centering
     \includegraphics[width=0.45\textwidth]{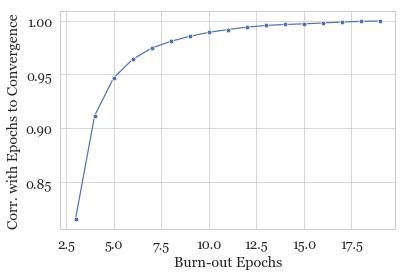}
     \caption{Pearson correlation coefficient of instance \variance on \winogrande between training dynamics when model is trained to convergence and when model is stopped early. 
     The high correlation indicates that training till convergence is not required to compute a good approximation of the training dynamics.
     } 
     \label{fig:burnout}
 \end{figure}

Moreover, it is possible to stop training early, or before the training converges for computing training dynamics.
This early \textit{burn-out} scheme results in \confidence and \variance measures which correlate well with \confidence and \variance (see Fig.~\ref{fig:burnout}).
For our experiments, we use later burn-outs corresponding to model convergence.

\subsection{Datasets}
\label{app:data}

This appendix provides further details on datasets.
We perform our experimental evaluation on four large datasets, each with at least 10K instances. 
Sizes of the different datasets are reported in \tabref{data_sizes}.
Instances in each of the original datasets are labeled by crowdworkers, whereas the OOD test sets are either manually or semi-automatically created.
The performance in each case is reported as accuracy.

\paragraph{\winogrande} 
This dataset contains a large scale crowd-sourced collection of Winograd schema challenge (\wsc \citealp{levesque2011winograd}) style questions.
Commonsense reasoning is required to select an entity from a pair of entities to complete a sentence.
Following \citet{sakaguchi2019winogrande}, we use the multiple choice architecture based on \roberta \cite{Liu2019RoBERTaAR}.
For OOD evaluation, we use the validation set from the original \wsc as provided under the SuperGLUE benchmark \cite{wang2019superglue}.
We used a rule-based method to convert \wsc validation and training data to the cloze-style format followed in \winogrande, removing all the repetitions included in the training data.
Figure~\ref{fig:winogrande} shows the data map for \winogrande. 
\swabha{Not sure how to explain the weird bump in \variance?}

\paragraph{\snli and \mnli} 
The task of natural language inference involves prediction of the relationship between a premise and hypothesis sentence pair.
The label determines whether the hypothesis entails, contradicts or is neutral to the premise.
We experiment with the Stanford natural language inference (\snli) dataset \cite{bowman-etal-2015-large} and its multi-genre counterpart, \mnli \cite{williams-etal-2018-broad}.\footnote{
For MultiNLI, we use the version released under the GLUE benchmark \cite{wang-etal-2018-glue}.}
Several challenge sets have been proposed to evaluate models OOD. 
As an OOD test set, we consider NLI diagnostics \cite{wang-etal-2018-glue} which contains a set of hand-crafted examples designed to demonstrate NLI model performance on several fine-grained semantic categories, such as lexical semantics, logical reasoning, predicate argument structure and commonsense knowledge.
In addition, we also report performance on the OOD mismatched \mnli validation set.
Figure~\ref{fig:multinli} shows the data map for \mnli.

\paragraph{\qnli} 
\citet{Rajpurkar2016SQuAD10} proposed the \squad dataset containing question and document pairs, where the answer to the question is a span in the document. 
The \qnli dataset, provided as part of the GLUE benchmark \cite{wang-etal-2018-glue} reformulates this as a sentence-level binary classification task. 
Here, the goal is to determine if a candidate sentence from the document contains the answer to the given question. 
As an OOD test set, we consider the \adversarial challenge set \cite{jia-liang-2017-adversarial} where distractor sentences are added to the document to confound the model.
We automatically convert this to the \qnli format.
Figure~\ref{fig:qnli} shows the data map for \qnli.

\begin{table}
\small
\centering
\begin{tabular}{lrrrr}
\toprule
& \multicolumn{3}{c}{In-dist.}  & \multicolumn{1}{c}{Out-of-dist.}  \\
\cmidrule(lr){2-4} \cmidrule(lr){5-5}
            & Train    &  Val.        & Test  & Test \\ 
\midrule
\winogrande &     40399     & 1268      &  -            & 424  \\ 
\snli       &     549368    & 9843      &  9825            & 1105   \\ 
\mnli       &     392703    & 9816      &     9833           & 1105  \\ 
\qnli       &     104744    & 5464      &     -           &  5324 \\ 
\bottomrule
\end{tabular}
\caption{
Dataset sizes. 
ID test set in \mnli is the mismatched validation set, which we did not use for validation, but as test.
We did not use the provided test sets in \winogrande and \qnli, rather report OOD performance for both cases.
}
\label{tab:data_sizes}
\end{table}

\subsection{Experimental Settings}
\label{app:experiments}

For each of our classifiers, we minimize cross entropy with the Adam optimizer \cite{kingma2014adam} following the AdamW learning rate schedule from \texttt{PyTorch}\footnote{\url{pytorch.org}}.
Each experiment is run with 3 random seeds and a learning rate\footnote{Learning rate is chosen using a log-uniform sampling strategy from the range (5e-6, 2e-5).} chosen using the \texttt{AllenTune} package \cite{dodge2019show}.
Initializations greatly affect performance, as noted in \citet{Dodge:2020}.
\winogrande and \snli \roberta-large models are trained for 6 epochs, and \mnli and \qnli are trained for 5 epochs each.
Each experiment is performed on a single Quadro RTX 8000 GPU.
Based on the available GPU memory, our experiments on all datasets use a batch size of 96, except for \winogrande, where a batch size of 64 is used.
Our implementation uses the \texttt{Huggingface Transformers} library \cite{Wolf2019HuggingFacesTS}.
For the active learning baselines, we train a acquisition model using \roberta-large on a randomly sampled 1\% subset of the full training set.
\subsection{\snli Qualitative Analysis}
\label{app:qualitative}

Qualitative samples from different regions of the \snli data map are provided in \tabref{more_qualitative}.
\todo{Add other datasets.}

\begin{table*}[t]
\centering
\small
\begin{tabular}{lp{5.8cm}p{4cm}cc}
\toprule 
  & \textbf{Premise} & \textbf{Hypothesis} & \textbf{Gold Label} & \textbf{Our Assessment}\\
\midrule 
\multirow{2}{*}{\rotatebox{90}{\challenging}}
& A mom is feeding two babies. 
& A mom is giving her children carrots to eat.
& \colorbox{blue!30!}{Contradiction$^-$} 
& Neutral
\\\cmidrule[0.01em](lr){2-5}
& Smiling woman in a blue apron standing in front of a pile of bags and boxes.
& The woman is wearing a red dress.
& Neutral 
&
\\\midrule[0.03em]
\multirow{7}{*}{\rotatebox{90}{\noisy}}
& Photographers take pictures of a girl sitting in a street.
& The photographer is taking a picture of a boy.
& \colorbox{blue!30!}{Entailment$^-$} 
& Contradiction
\\\cmidrule[0.01em](lr){2-5}
& A group of men in a blue car driving on the track.
& One woman is driving the blue car.
& \colorbox{blue!30!}{Entailment$^-$}
& Contradiction
\\\cmidrule[0.01em](lr){2-5}
& Pedestrians walking down the street passing The Temple Bar.
& The pedestrians are outside.
& \colorbox{blue!30!}{Contradiction$^-$} 
& Entailment
\\\midrule[0.03em]
\multirow{5}{*}{\rotatebox{90}{\easy}}
& Four musicians play their instruments on the street while a young man on a bike stands by to listen.
&  a kid in a car goes through a drive thru	
& Contradiction
&
\\\cmidrule[0.01em](lr){2-5}
& A girl sits with excavating tools examining a rock.
& Two men writing a draft of a speech.	
& Contradiction
&
 \\
\bottomrule
\end{tabular}
\caption{
Examples from \snli belonging to different regions in the data map.
Cases where authors disagree with the gold standard are highlighted in \colorbox{blue!30!}{blue$^-$}.
} 
\label{tab:more_qualitative}
\end{table*}
\section{Additional Results}
\label{app:addition_results}

Results on the \snli validation set are provided in \tabref{snli_val}.

\begin{table}
\small
\centering
\begin{tabular}{llc}
\toprule
\cmidrule(lr){3-3}  & & \snli Val. (In-dist.)   \\ 
\midrule
&100\% train     & 93.1  \\ 
\midrule[0.03em]
\multirow{3}{*}{\rotatebox{90}{33\% train}} & \textsl{random}          & 92.1  \\ 
\cmidrule[0.03em]{2-3}
& \noisy     & 92.6  \\
& \challenging     & 92.9  \\
\bottomrule
\end{tabular}
\caption{
\snli validation performance comparing different selection methods.
Reported numbers are the best of 3 runs across different seeds.
}
\label{tab:snli_val}
\end{table}

\subsection{Training Dynamics vs. Dropout}
\label{app:training-dynamics-vs-dropout}

To empirically test the hypothesis that \confidence and \variance from the training dynamics respectively quantify intrinsic and model uncertainty, we compare \confidence and \variance against an established method of capturing intrinsic and model uncertainty from the literature based on dropout \cite{srivastava2014dropout}. Dropout can be seen as variational Bayesian inference \cite{gal2016dropout}, with predictions from different dropout masks corresponding to predictions sampled from the posterior. Thus, \confidence and \variance computed from sampled dropout predictions measure the average and standard deviation of the gold label's probability under the posterior---quantifying the intrinsic and model uncertainty in a principled way.

\begin{figure}[tbh]
     \centering
     \includegraphics[width=1.1\columnwidth]{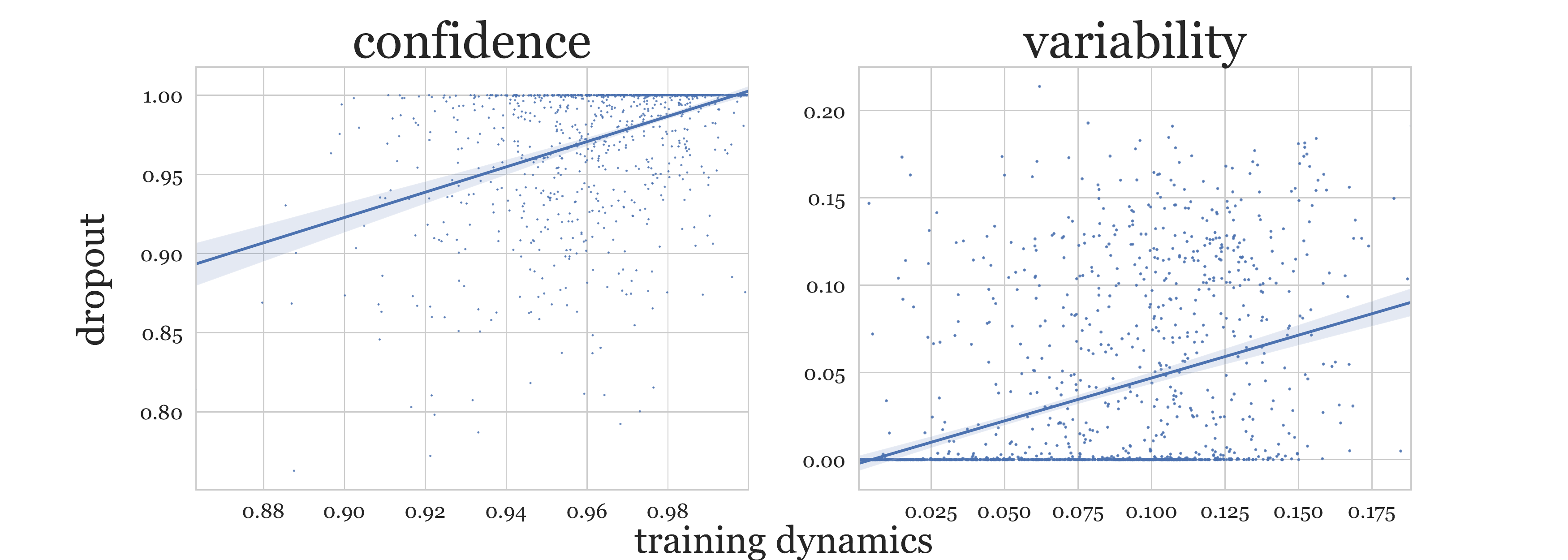}
     \caption{
     The \confidence (left) and \variance (right) from sampled dropout predictions correlate positively with those from the training dynamics on \winogrande (dev.~set). Shaded regions are bootstrapped 95\% confidence intervals for the regression line.
     }
     \label{fig:dropout-vs-training-dynamics}
\end{figure}

We computed \confidence and \variance from both training dynamics and dropout on \winogrande's development set.\footnote{To compute training dynamics, we trained a model on the combined training and development sets for \winogrande. In contrast, the dropout model was trained only on \winogrande's training set then run on development, to avoid over-fitting and provide higher quality uncertainty estimates.}
Figure~\ref{fig:dropout-vs-training-dynamics} visualizes a regression analysis of the relationship between \confidence and \variance from training dynamics and dropout. \confidence from training dynamics and dropout correlate between 0.450 and 0.452 for Pearson's $r$ at 95\% confidence. Likewise, \variance from training dynamics and dropout share a Pearson's $r$ from 0.390 to 0.393 at 95\% confidence. 
Thus, the training dynamics empirically demonstrate a positive, predictive relationship with these first-principles estimates of the intrinsic and model uncertainty.
Compared to dropout, however, training dynamics have the pragmatic advantage that all information required to calculate them is already available from training, without additional work or computation.

\section{Additional Data Maps}
\label{app:all_data_maps}

\begin{figure*}
\centering
\begin{subfigure}[b]{\textwidth}
\centering
   \includegraphics[width=\textwidth]{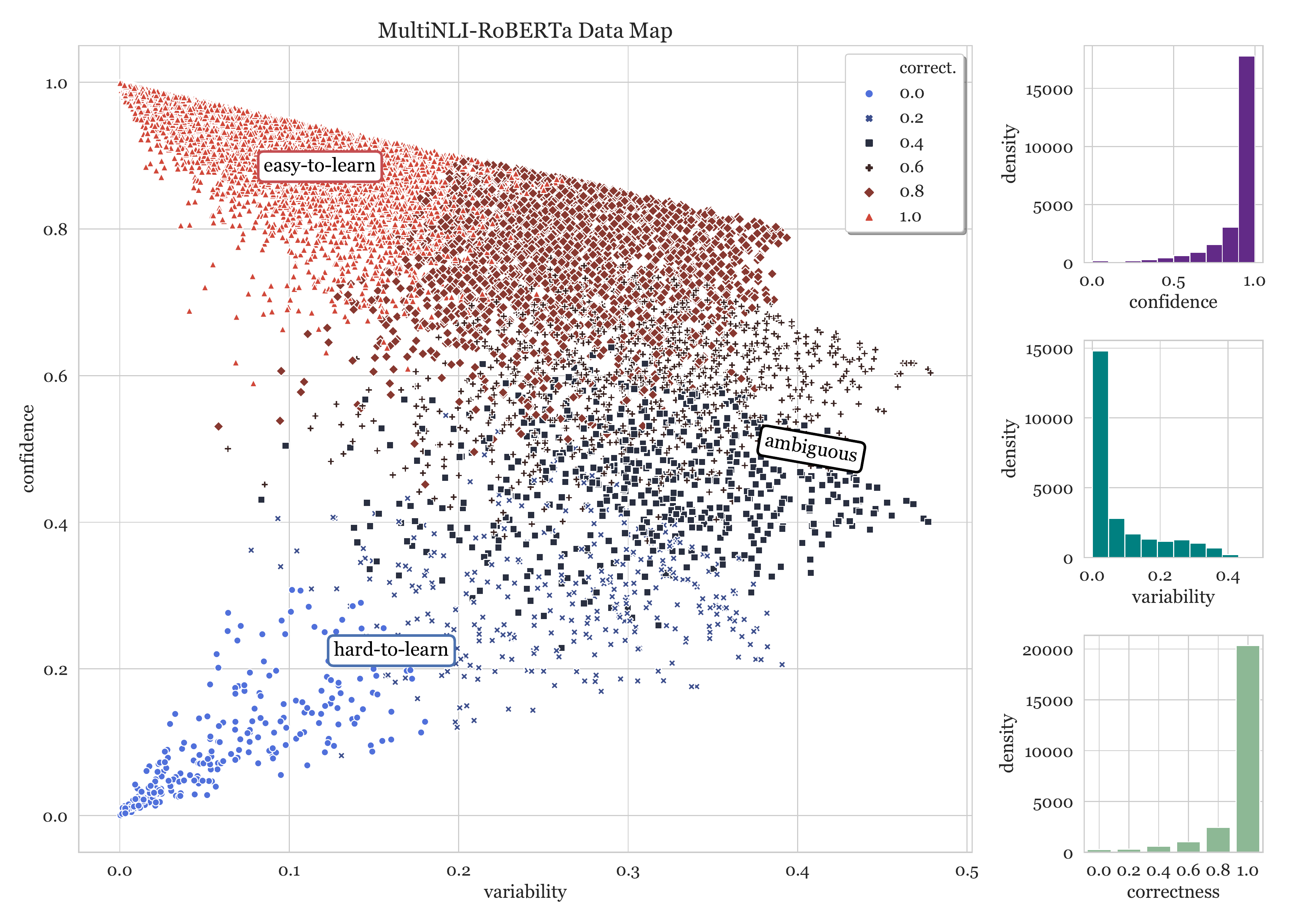}
     \caption{Data Map for \mnli \cite{williams-etal-2018-broad} and density plots for different measures based on training dynamics (below). 
     For clarity we only use 50K random samples from \mnli in the scatter plot.
     Trends are very similar to \snli, even though \mnli contains samples from diverse genres.
     } 
   \label{fig:multinli} 
\end{subfigure}
\begin{subfigure}[b]{\textwidth}
\centering
   \includegraphics[width=\textwidth]{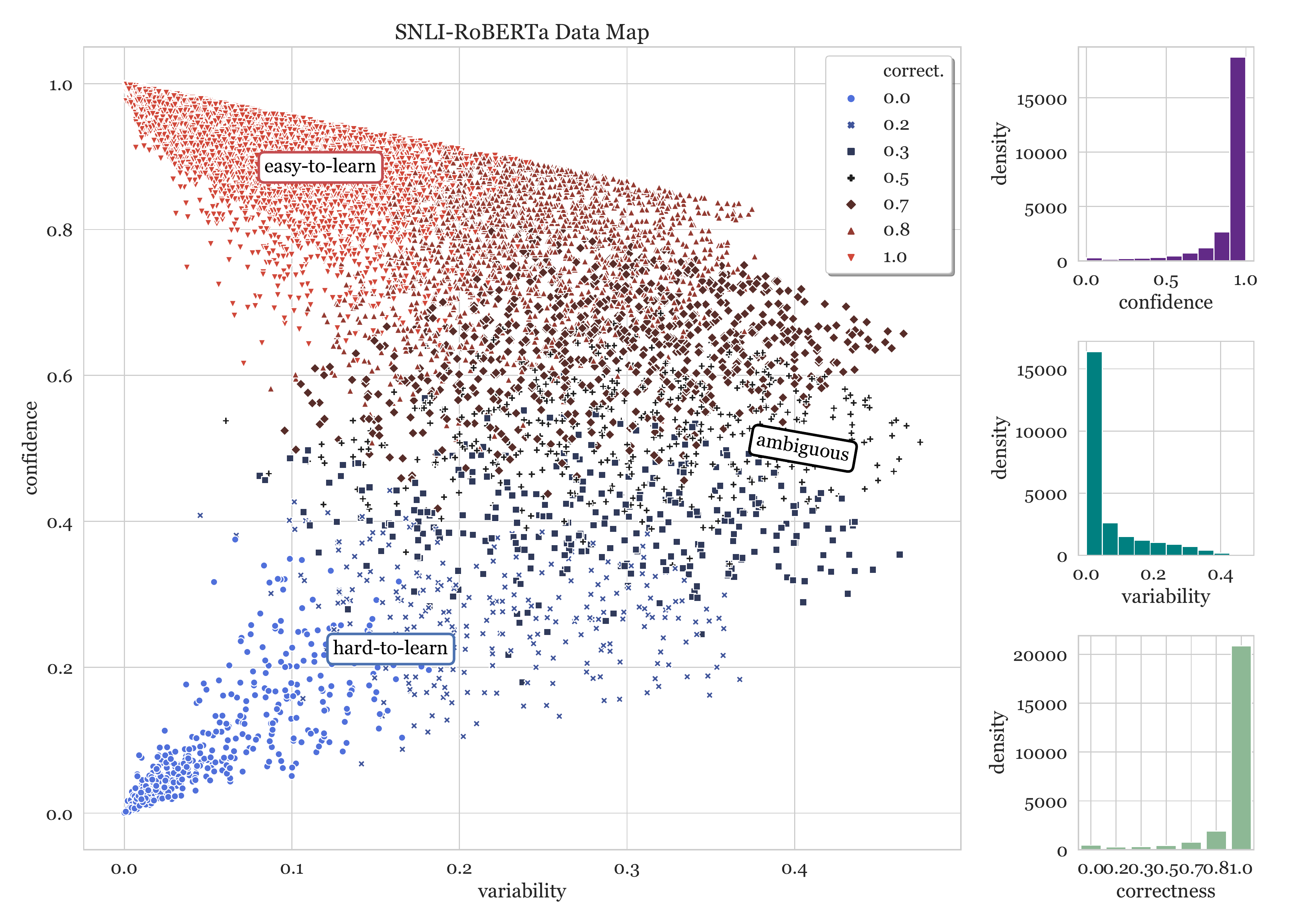}
     \caption{(Left) Data Map for \snli, same as \figref{snli}, provided here in greater relief again for comparison with other datasets. \snli is larger than all other datasets, and thus has a higher density of \easy examples. 
     (Right) Densities of the above statistics across the entire dataset; examples which are \easy (for \roberta) form the vast majority of \snli.
     } 
        \label{fig:snli2}
\end{subfigure}
\caption{Data maps for *NLI datasets; each data map plots 25K instances, for clarity.}
\label{fig:data_maps3}
\end{figure*}

All the data maps have been provided in \figref{data_maps3} and \figref{data_maps2}.

\subsection{Effect of Encoder in building Data Maps}
\label{sec:encoder}

\begin{figure*}
\centering
\begin{subfigure}[b]{\textwidth}
\centering
   \includegraphics[width=\textwidth]{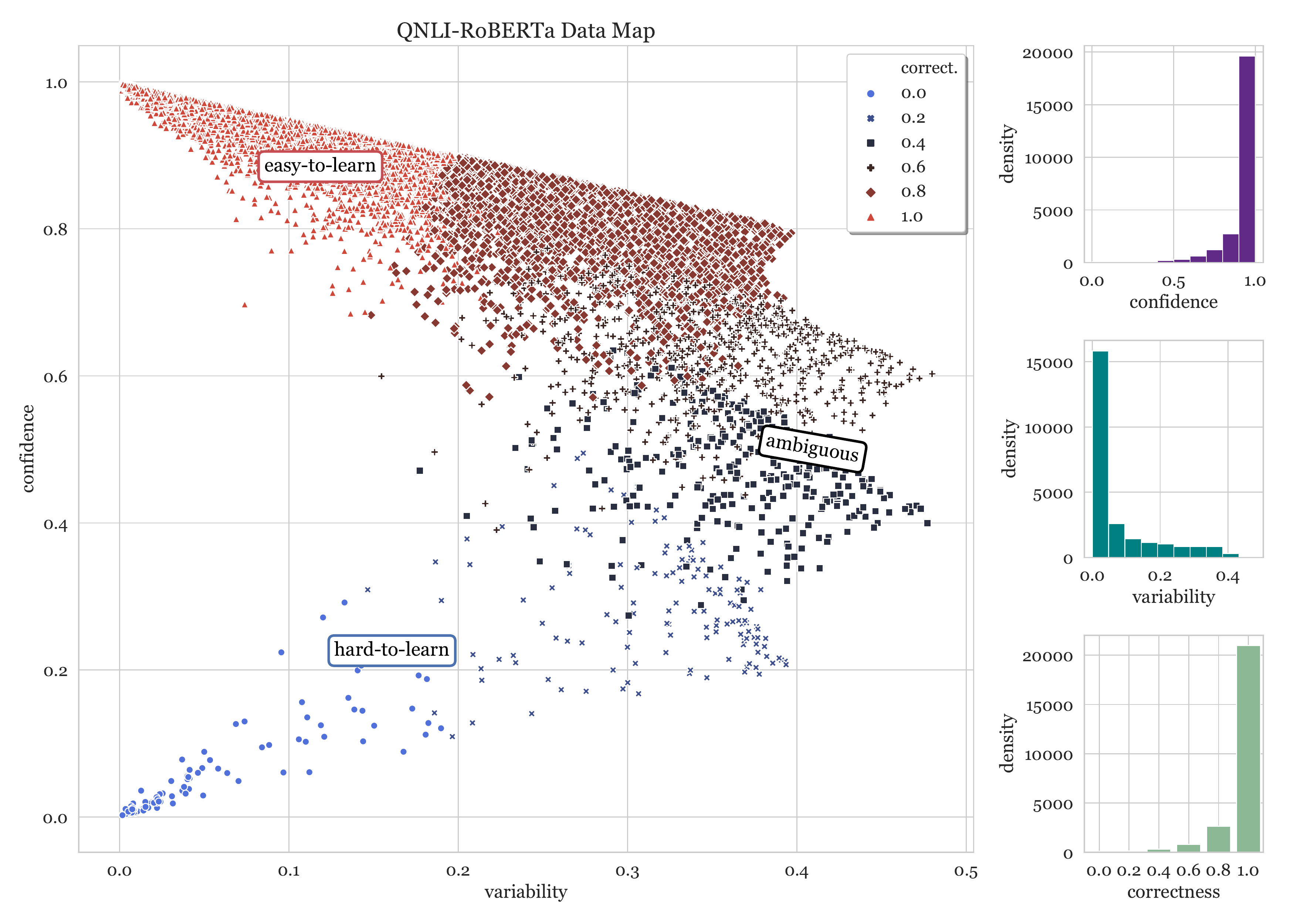}
     \caption{
     Data map for the sentence-level \squad dataset, \qnli (left) and density plots for different measures based on training dynamics (right). 
     Unlike other datasets, \qnli has fewer instances with low \variance and \confidence close to 0.5.
     } 
    \label{fig:qnli}
\end{subfigure}
\begin{subfigure}[b]{\textwidth}
\includegraphics[width=\textwidth]{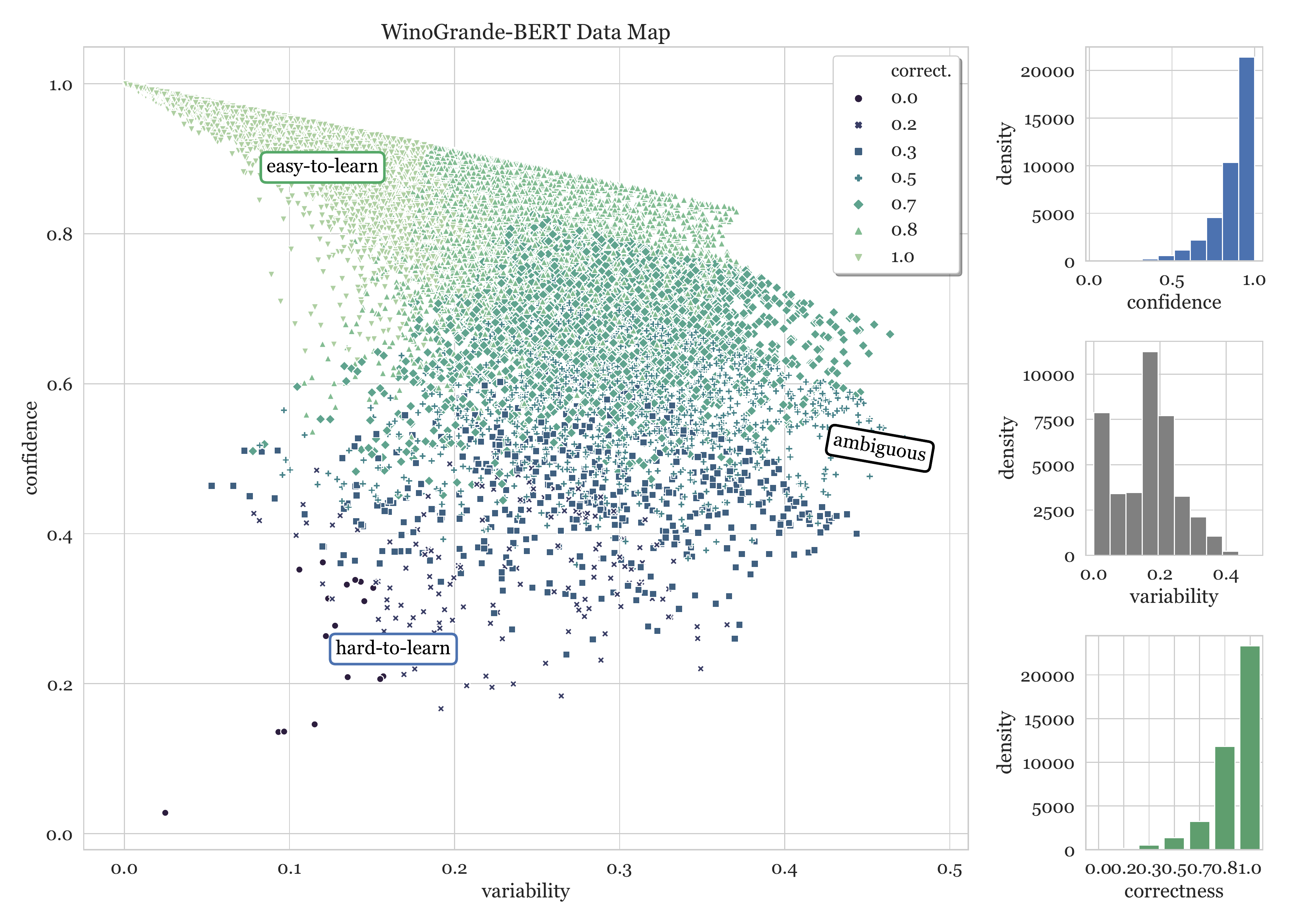}
     \caption{ 
     Data map for \winogrande built based on a \textbf{\bert}-large \cite{Devlin:2019} model. 
     While similar regions can be seen as a \winogrande-\roberta data map (\figref{winogrande}), the densities of different regions can be different.
     Moreover the same instances might be mapped to different regions across maps.
     } 
  \label{fig:bert_winogrande} 
\end{subfigure}
\caption{Additional data maps, each plotting 25K instances, for clarity.}
\label{fig:data_maps2}
\end{figure*}

While training dynamics are inherently model dependent, data maps can be built for any model, and might reveal similar structures.
Since models can be of varying capacities with respect to a task or dataset, instances might receive different co-ordinates on data maps built based on different models.
For instance, \bert is known to be worse at reasoning than \roberta \cite{sakaguchi2019winogrande,Talmor2019oLMpicsO}, and \roberta being a larger model is likely very sample efficient \cite{kaplan2020scaling}.
However, the overall structure of data maps based on different models remains the same; \figref{bert_winogrande} shows the data map built for \winogrande using a \bert-large classifier.

\begin{figure*}
    \centering
    \begin{subfigure}[b]{\textwidth}
      \includegraphics[width=\textwidth]{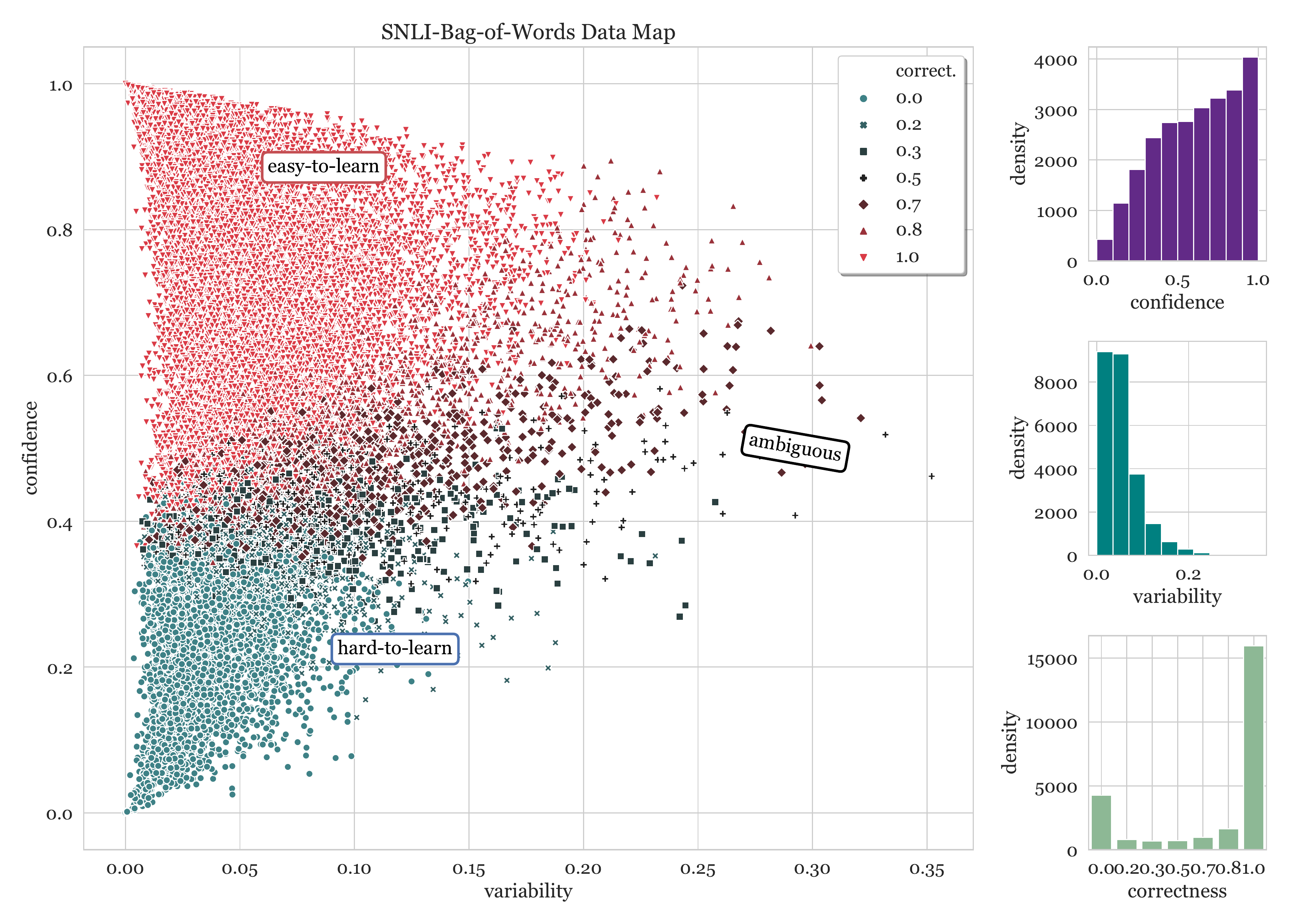}
    \end{subfigure}
    \begin{subfigure}[b]{\textwidth}
       \includegraphics[width=\textwidth]{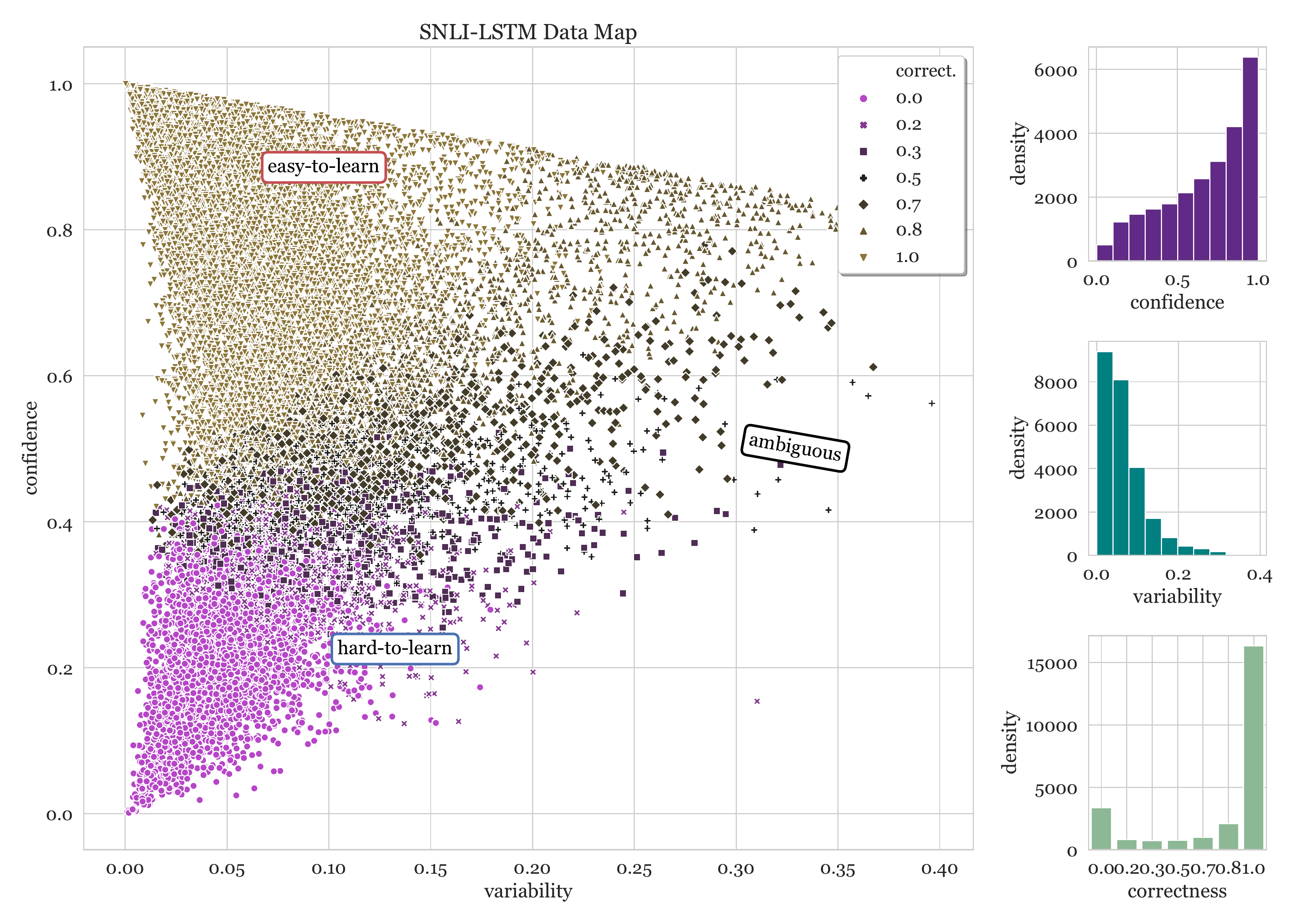}
    \end{subfigure}
    \caption{Data maps for SNLI based on non-\roberta (and weaker) architectures---bag of words (BoW; Top) and LSTMs (Bottom).
    Although these maps exhibit bell-shaped curves, similar to the \roberta data map for SNLI in \ref{fig:snli2}, the curvature is somewhat smaller. 
    The spread of the data is larger across the regions, which are not as distinct as in the \roberta data map.
    These shapes could be attributed to these architectures being weaker (and hence unable to overfit to data) than those involving representations from large, pretrained language models.
    Each data map plots 25K instances, for clarity, and are best viewed enlarged.
    }
    \label{fig:lstm_bow_snli}
\end{figure*}

\begin{figure*}
    \centering

    \begin{subfigure}[b]{\textwidth}
       \includegraphics[width=\textwidth]{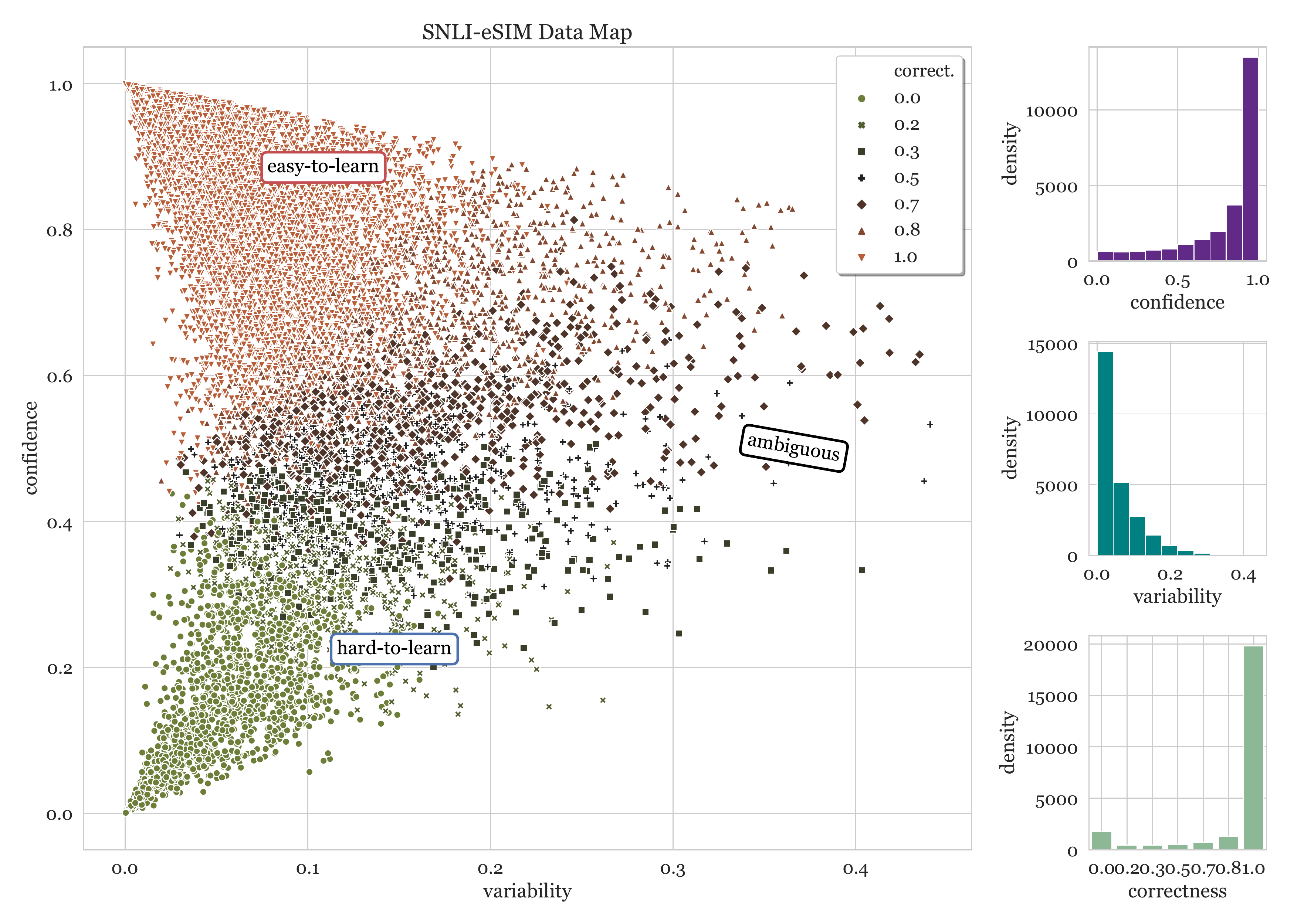}
    \end{subfigure}
    \begin{subfigure}[b]{\textwidth}
       \includegraphics[width=\textwidth]{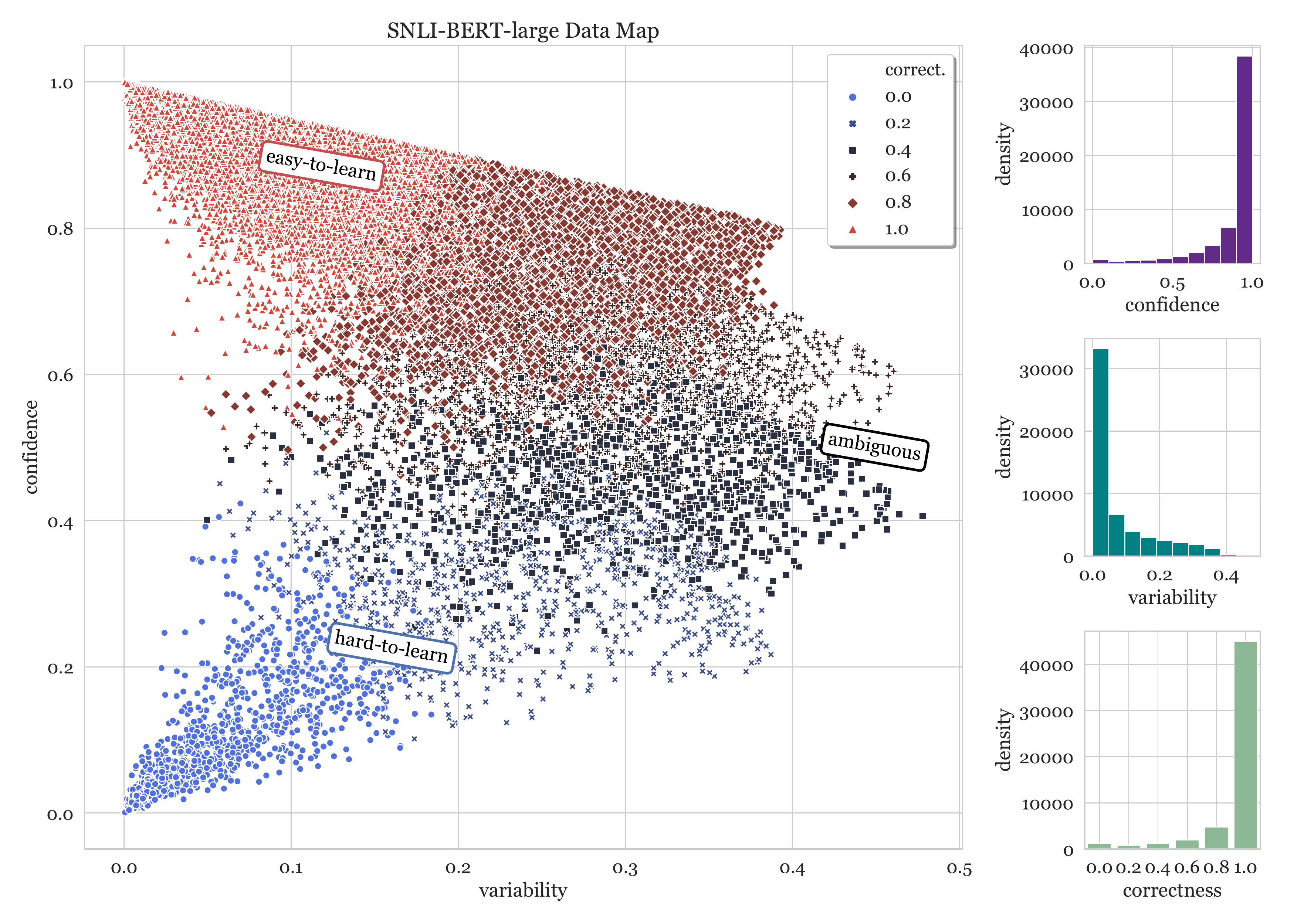}
    \end{subfigure}
    \caption{Data maps for SNLI based on other (weaker) architectures---bag of words eSim~\cite{chen-etal-2017-enhanced} (Top) and BERT-large (Bottom).
    Although these maps exhibit bell-shaped curves, similar to the \roberta data map for SNLI in \ref{fig:snli2}, the curvature is somewhat smaller for eSIM. 
    The spread of the data is larger across the regions, which are not as distinct as in the \roberta data map.
    Each data map plots 25K instances, for clarity, and are best viewed enlarged.
    }
    \label{fig:bert_esim_snli}
\end{figure*}

Four different architectures for the SNLI dataset are compared in \figref{lstm_bow_snli} and \figref{bert_esim_snli}.

\end{document}